\documentclass[10pt,twocolumn,letterpaper]{article}

\usepackage{cvpr}              

\usepackage[accsupp]{axessibility}
\usepackage{graphicx}
\usepackage{amsmath}
\usepackage{amssymb}
\usepackage{booktabs}
\usepackage{verbatim}
\usepackage{times}
\usepackage{epsfig}
\usepackage{makecell}
\usepackage{dsfont}
\usepackage{enumitem}
\usepackage{caption}
\usepackage{arydshln}
\usepackage{stmaryrd}
\usepackage[normalem]{ulem}
\usepackage[ruled,vlined]{algorithm2e}
\usepackage{UserCommands}

\usepackage[pagebackref,colorlinks]{hyperref}

\usepackage[capitalize]{cleveref}
\crefname{section}{Sec.}{Secs.}
\Crefname{section}{Section}{Sections}
\Crefname{table}{Table}{Tables}
\crefname{table}{Tab.}{Tabs.}

\def\cvprPaperID{3888} 
\def\confName{CVPR}
\def\confYear{2022}

\begin{document}

\title{TransFusion: Robust LiDAR-Camera Fusion for 3D Object Detection with Transformers 
} 

\author{Xuyang Bai$^{1}$\hspace{0.15cm} Zeyu Hu$^{1}$\hspace{0.15cm} Xinge Zhu$^{2}$\hspace{0.15cm} Qingqiu Huang$^{2}$\hspace{0.15cm} Yilun Chen$^{2}$\hspace{0.15cm} Hongbo Fu$^{3}$\hspace{0.15cm}  Chiew-Lan Tai$^{1}$ \\
\normalsize $^1$Hong Kong University of Science and Technology \hspace{0.7cm} $^2$ADS, IAS BU, Huawei \hspace{0.7cm} $^3$City University of Hong Kong\\

}

\maketitle

%



\begin{abstract}
LiDAR and camera 
are two important sensors for 3D object detection in autonomous driving. 
Despite the increasing popularity of 
{sensor fusion} in this field, the robustness {against} 
inferior image conditions, e.g., 
bad illumination and sensor misalignment, {is under-explored.}
Existing fusion methods
{are easily affected by such conditions, mainly due to 
{a hard association}
of LiDAR points and image pixels, established by calibration matrices.}

{We propose \Name, a robust solution to LiDAR-camera fusion}
with a \emph{soft-association} mechanism {to} 
handle inferior image conditions.
Specifically, our \Name~consists of convolutional backbones and a detection head based on {a} transformer decoder. The first layer of the decoder 
predicts initial bounding boxes from {a} LiDAR point cloud using a sparse set of object queries, and its second decoder layer adaptively fuses {the} object queries with useful image features, leveraging both spatial and contextual relationship{s}. The attention mechanism of the transformer enables our model to adaptively determine where and what information should be taken from the image, leading to a robust and effective fusion strategy. We additionally design an image-guided query initialization strategy to deal with objects that are difficult to {detect} 
in point clouds.
\Name~achieves state-of-the-art performance on {large-scale} 
datasets. We provide extensive experiments to demonstrate {its} 
robustness against {degenerated image quality}
and calibration errors.
We also extend the proposed method to {the} 3D tracking task and {achieve} the \textbf{1st} place in the leaderboard of nuScenes tracking{, showing} 
its effectiveness and generalization capability. [\href{https://github.com/XuyangBai/TransFusion/}{code release}]

\end{abstract}



 \vspace{-0.1cm}
\section{Introduction}
 \vspace{-0.1cm}
{As one of the fundamental tasks in self-driving}, 3D object detection aims {to localize} 
a set of objects in 3D space and {recognize}
their categories. 
Thanks to the accurate depth information provided by LiDAR, early works such as VoxelNet~\cite{Zhou2018VoxelNetEL} and PointPillar~\cite{Lang2019PointPillarsFE} achieve reasonably {good} results using only point clouds as input. However, these LiDAR-only
methods are generally surpassed by the methods using both LiDAR and camera data on large-scale datasets with sparser point clouds, such as nuScenes~\cite{Caesar2020nuScenesAM} and Waymo~\cite{Sun2020ScalabilityIP}.
{LiDAR-only methods are surely insufficient for robust 3D detection due to the sparsity of point clouds.}
For example, 
small or {distant} objects 
are difficult to {detect} 
in 
LiDAR modality. {In contrast, such} 
objects are {still} clearly visible and distinguishable in 
high-resolution 
images.
The complementary roles of point clouds and images motivate {researchers} 
to design {detectors} 
utilizing the best of {the} two worlds, {i.e., multi-modal detectors}. 

Existing LiDAR-camera fusion methods roughly fall into three categories: \emph{result-level}, \emph{proposal-level}, and \emph{point-level}. The result-level methods, including FPointNet~\cite{Qi2018FrustumPF} and RoarNet~\cite{Shin2019RoarNetAR}, 
 use off-the-shelf 2D detectors to seed 
 3D proposals, followed by a PointNet~\cite{Qi2017PointNetDL} {for object localization}. 
 The proposal-level fusion methods, including MV3D~\cite{Chen2017Multiview3O} and AVOD~\cite{Ku2018Joint3P}, perform fusion at the region proposal level by applying RoIPool~\cite{Ren2015FasterRT} in each modality for shared proposals. 
These coarse-grained fusion methods show unsatisfactory results since rectangular 
region{s} of interest~(RoI)
usually contain lots of background noise.
Recently, a majority of approaches have {tried} 
to do point-level fusion and achieved promising results. They first find a {hard association}
between LiDAR points and image pixels {based on calibration matrices}, {and} then augment LiDAR features with the segmentation scores~\cite{Vora2020PointPaintingSF, Xu2021FusionPaintingMF} or CNN features~\cite{Sindagi2019MVXNetMV, Meyer2019SensorFF, Huang2020EPNetEP, Wang2021PointAugmentingCA, Zhang2020MultiModalityCA} of {the} associated pixels through point-wise concatenation.
Similarly, \cite{Liang2019MultiTaskMF, Liang2018DeepCF, Xie2020PIRCNNAE, Yoo20203DCVFGJ} first project 
{a} point cloud onto {the} bird's eye view~(BEV) plane and then fuse the image features with {the} BEV pixels. 

\begin{figure}[t]
\vspace{-0.5cm}
\setlength{\abovecaptionskip}{0.0cm}
\setlength{\belowcaptionskip}{-0.45cm}
\centering
    \includegraphics[width=8cm]{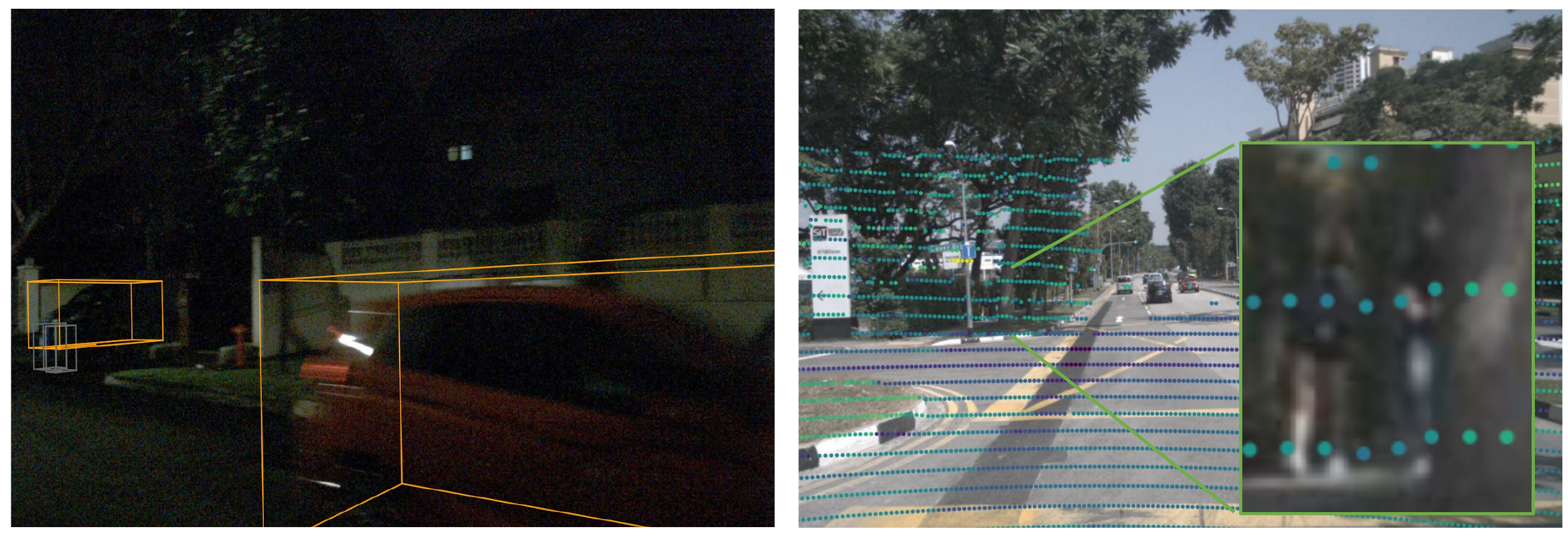}
    \caption{Left: An example of bad illumination conditions. Right: Due to the sparsity of point clouds, the hard-association based fusion methods waste many image features and are sensitive to sensor calibration, since the projected points may fall outside {objects} 
    due to a small calibration error.}
    \label{fig:lidar_vs_img}
\end{figure}

 Despite the impressive improvements, these point-level fusion methods suffer from two major problems, as shown in Fig.~\ref{fig:lidar_vs_img}. First,
{they simply fuse} the LiDAR features and image features 
through element-wise addition or concatenation, {and thus their} 
performance degrades {seriously} 
with low-quality image features, e.g., {images in bad illumination conditions.}
Second, finding the hard association between sparse LiDAR points and dense image pixels not only wastes many image features with rich semantic information, but also {heavily relies on} 
high-quality calibration between two sensors, which is usually hard to acquire due to the inherent spatial-temporal misalignment~\cite{Zhao2021LIFSegLA}. 
 
{To address the shortcomings of the previous fusion approaches,}
we introduce an effective and robust multi-modal detection framework in this paper. 
Our 
{key} 
idea is to reposition the focus of {the} fusion process, from hard-association to soft-association, 
leading to the robustness against degenerated image 
{quality} and sensor misalignment.

Specifically, we design a sequential fusion method {that} uses two transformer decoder layers {as the detection head.}
{To our best knowledge, we are the first to use transformer for LiDAR-camera 3D detection.}
{Our first decoder layer leverages a sparse set of object queries to produce initial bounding boxes from LiDAR features.}
Unlike \prevqueryinit~object queries in 2D~\cite{carion2020endtoend, Sun2020SparseRE}, we 
make the object queries \emph{\ourqueryinit} and \emph{category-aware} so that the queries are enriched with better position and category {information}. 
Next, {the second transformer decoder layer adaptively fuses} 
object queries with useful image features associated by {spatial and contextual relationships}.  
We leverage a locality inductive bias by spatially constraining the cross attention around the initial bounding boxes to help the network better visit the related positions. 
Our fusion module not only provides rich semantic information to object queries, but also is more robust to 
inferior image conditions since the association between LiDAR points and image pixels are established in a soft and adaptive way. 
{Finally, to handle objects that are difficult to detect in point clouds, we introduce an \secondmodule~module to involve image guidance on the query initialization stage.}
{Overall, the corporation of these components significantly improves the effectiveness and robustness of our LiDAR-camera 3D detector.}
To summarize, our contributions are fourfold:
\begin{enumerate}[itemsep=-1mm]
\vspace{-0.2cm}
\item[1.] Our studies investigate the inherent difficulties of {LiDAR-camera} 
fusion and reveal a crucial aspect to 
{robust} fusion, namely, the soft-association mechanism.
\item[2.] We propose a novel transformer-based LiDAR-camera fusion model for 3D detection, which performs fine-grained fusion in an attentive manner and shows superior robustness against degenerated image {quality} and sensor misalignment.
\item[3.] {We introduce} several simple yet effective 
adjustments {for} 
object queries 
to boost the quality of initial bounding box predictions for image fusion.
An \secondmodule~module is also designed to handle objects that are hard to detect in point clouds.
\item[4.] We achieve the state-of-the-art 3D detection performance on nuScenes {and competitive results on Waymo}.
{We also extend our model to the 3D tracking task and achieve {the} \textbf{1st place} in the leaderboard of {the} nuScenes tracking challenge.}

\end{enumerate}
\section{Related Work}

\noindent\textbf{LiDAR-only 3D Detection} aims to predict 3D bounding boxes {of objects in} given point clouds~\cite{Yang2018PIXORR3, Qi2019DeepHV, Qi2020ImVoteNetB3, Vora2020PointPaintingSF, Zhou2019EndtoEndMF, Chen2020EveryVC, Shi2021FromPT, Zhu2019ClassbalancedGA, Zhu2020SSNSS, Chen2020ObjectAH}. Due to {the} 
unordered, irregular nature {of point clouds}, many 3D detectors first project {them} 
onto {a} regular grid such as 3D voxels~\cite{Zhou2018VoxelNetEL, Yan2018SECONDSE}, pillars~\cite{Lang2019PointPillarsFE} or range images~\cite{Sun2021RSNRS, Fan2021RangeDetID}. After that, standard 2D or 3D convolutions are used to compute the features in {the} BEV plane, where objects are naturally separated{, with their} 
physical sizes 
preserved. 
Other works~\cite{Shi2019PointRCNN3O, Yang20203DSSDP3, Yang2019STDS3, Shi2020PVRCNNPF} directly {operate} 
on raw point clouds without quantization.
{The mainstream of 3D detection head is based on anchor boxes~\cite{Lang2019PointPillarsFE, Zhou2018VoxelNetEL} following the 2D counterparts, while
\cite{Yin2020Centerbased3O, Wang2020PillarbasedOD} adopt 
a center-based representation for 3D objects, largely simplifying the 3D detection pipeline}. 
{Despite the popularity of adopting {the} transformer architecture {as a} 
detection head in 2D~\cite{carion2020endtoend}, 3D detection models for outdoor scenarios {mostly} utilize {the} transformer for feature extraction~\cite{Pan20203DOD, Mao2021VoxelTF, Sheng2021Improving3O}.}
However, the attention operation in each {transformer} layer {requires {a} computation complexity of $\mathcal{O}(N^2)$ for $N$ points,} %
{{requiring} 
a carefully designed memory reduction operation 
when handling}
LiDAR point clouds with millions 
of points per frame. In contrast, our model retains {an} 
efficient convolution backbone for feature extraction 
and leverage{s} {a} transformer decoder {with a small set of object queries as} {the} detection head, {making the computation cost manageable}. {The concurrent works~\cite{Liu2021GroupFree3O, Liu2021SuppressandRefineFF, misra20213detr} adopt transformer as a detection head but focus on indoor scenarios {and extending these methods to outdoor scenes is non-trivial.}}

\noindent\textbf{LiDAR-Camera
3D Detection} has gained increasing attention due to the complementary roles of point clouds and images. Early works~\cite{Qi2018FrustumPF, Shin2019RoarNetAR, Chen2017Multiview3O} adopt result-level or proposal-level fusion, where the fusion granularity is too coarse to release the full potential of two modalities. Since PointPainting~\cite{Vora2020PointPaintingSF} was proposed, the point-level fusion methods~\cite{Sindagi2019MVXNetMV, Huang2020EPNetEP, Wang2021PointAugmentingCA} {have} 
shown great advantages and promising results. However, such methods are easily affected by the sensor misalignment due to the {hard association} 
between points and pixels established by calibration matrices. Moreover, the simple point-wise concatenation ignores the {quality of} real data 
and {contextual} 
relationships between two modalities, {and thus} 
leads to degraded performance when the image features {are} 
defective. In our work, we explore a more robust and effective fusion mechanism to mitigate these limitations {during LiDAR-camera fusion}. 

\begin{figure*}[tbh]
\centering
 \vspace{-0.6cm}
 \setlength{\abovecaptionskip}{0.1cm}
 \setlength{\belowcaptionskip}{-0.25cm}
    \includegraphics[width=17cm]{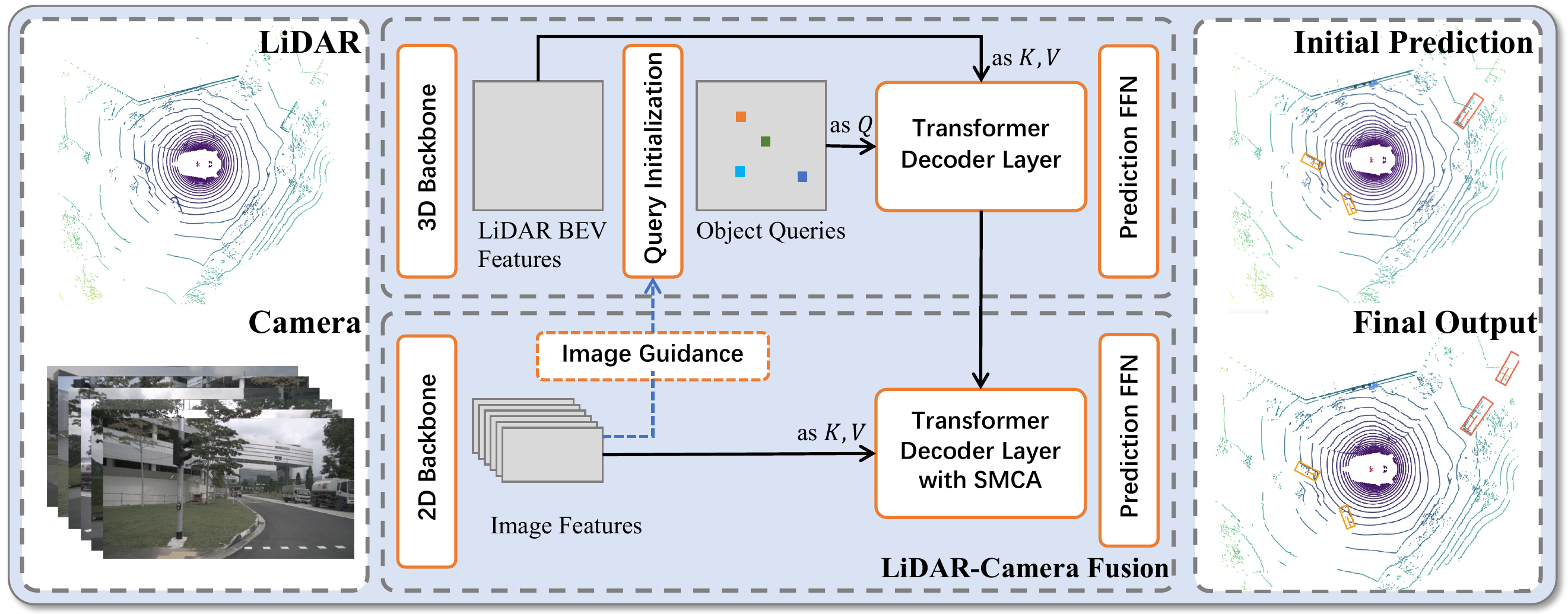}
    \caption{	
    Overall pipeline of \Name. 
Our model relies on standard 3D and 2D backbones to extract LiDAR BEV feature map and image feature map. 
    Our detection head consists of two transformer decoder layers sequentially: (1) The first layer produces initial 3D bounding boxes using a sparse set of object queries, 
    initialized in a \ourqueryinit~and category-aware manner. (2) The second layer attentively {associates and fuses} 
    the object queries~(with initial predictions) from the first stage with the image features, producing rich texture and color cues for better detection results. A spatially modulated cross attention~(SMCA) mechanism is introduced to involve a locality inductive bias and help {the} network better attend to the related image regions. We additionally propose an \secondmodule~strategy to involve image guidance on LiDAR BEV. This strategy 
    helps {produce} 
    object queries that are difficult to detect in the sparse LiDAR point clouds. 
   	}
    \label{fig:pipeline}
\end{figure*}

\section{Methodology}
	
In this section, we present the proposed method \Name~for LiDAR-camera 
3D object detection. As shown in Fig.~\ref{fig:pipeline}, given {a} LiDAR BEV feature map and {an} image feature map from convolutional backbones,
{our transformer-based detection head first decodes object queries into initial bounding box predictions using the LiDAR information, and then performs LiDAR-camera fusion by attentively fusing object queries with useful image features.}
Below we will first provide the preliminary knowledge about a transformer architecture for detection and then present the detail of \Name.





\subsection{Preliminary: Transformer for {2D} Detection}
Transformer~\cite{Vaswani2017AttentionIA} has been widely used for 2D object detection~\cite{Zhu2021DeformableDD, Sun2020SparseRE, Gao2021FastCO, Yao2021EfficientDI} since DETR~\cite{carion2020endtoend} was proposed. 
DETR uses {a} CNN backbone to extract image features and {a} transformer 
architecture to convert a small set of learned embeddings~(called object queries) into a set of predictions. The follow{-}up works~\cite{Zhu2021DeformableDD, Sun2020SparseRE, Yao2021EfficientDI} further equip the object queries with positional information~\footnote{Slightly different concepts might be introduced, e.g., reference points in Deformable-DETR~\cite{Zhu2021DeformableDD} and proposal boxes in Sparse-RCNN~\cite{Sun2020SparseRE}.}. 
The final predictions of boxes are the relative offsets w.r.t. the query positions to reduce 
optimization difficulty. We refer readers to the original papers~\cite{carion2020endtoend, Zhu2021DeformableDD} for more details. 
{In our work,}
each object query contains {a} query position providing the localization of the object and a query feature
 encoding instance information, such as the {box's} 
size, orientation, etc. 

\subsection{Query Initialization}
\label{subsubsec: query_initialize}
\noindent\textbf{Input-dependent.} The query positions in {the seminal} 
works~\cite{carion2020endtoend, Zhu2021DeformableDD, Sun2020SparseRE} are randomly generated or learned as network parameters{, regardless of the input {data}.}
Such \prevqueryinit~
query positions will take extra stages~(decoder layers) for 
{their models~\cite{carion2020endtoend, Zhu2021DeformableDD}}
to learn the moving process towards the real object centers. Recently, it has been {observed} in 2D object detection~\cite{Yao2021EfficientDI} that with a better initialization of object queries, the gap between 1-layer structure and 6-layer structure could be bridged. 
{Inspired by this observation,}
we propose an \ourqueryinit~initialization strategy based on a center heatmap to achieve competitive performance using only one decoder layer. 

{Specifically}, given a $d$ 
dimensional LiDAR BEV feature map $F_{L} \in \mathbb{R}^{X\times Y\times d}$, we first predict a class-specific heatmap $\hat S \in \mathbb{R}^{X\times Y\times K}$, where $X\times Y$ {describes} 
the size of {the} BEV feature map and $K$ is the number of categories. Then we regard the heatmap as $X\times Y\times K$ object candidates and select the top-$N$ candidates {for all the categories} as our initial object queries. 
To avoid spatially too closed queries, following~\cite{Zhou2019ObjectsAP},
we select {the} local maximum elements {as our object queries},
whose values are greater than or equal to {their} 8-connected neighbors. Otherwise{,} a large number of queries are needed to cover the BEV plane. 
The positions and features of the selected candidates are used to initialize the query positions and query features. In this way, our initial object queries will locate at or close to the potential object centers, eliminating the need of multiple decoder layers~\cite{misra20213detr, Liu2021GroupFree3O, Wang2021ObjectD3} to refine the locations.

\noindent\textbf{Category-aware.}  
Unlike the{ir} 2D projections on the image plane, the objects on the BEV plane are all in 
absolute scale and has small scale variance 
among the same categories. To leverage such properties for better multi-class detection, we make the object queries category-aware by equipping each query with a category embedding. Specifically, using the category of each selected candidate~(e.g. $\hat S_{ijk}$ {belonging} 
to the $k$-th category), we element-wisely sum the query feature with a category embedding produced by linearly projecting the one-hot category vector into a $\mathbb{R}^d$ vector. The category embedding brings benefits 
{in} two aspects: on {the} one hand, it serves as a useful side information when modelling the object-object relations in {the self-attention 
modules}
and the object-context relations in {the cross-attention modules.}
On the other hand, during prediction, it could deliver valuable prior knowledge of the object, making the network {focus} 
on intra-category variance and {thus} benefiting the property prediction. 





\subsection{Transformer Decoder and FFN} 


\noindent The decoder layer follows the {design} of DETR~\cite{misra20213detr} 
and the detailed architecture is provided in the supplementary Sec.~\ref{supp:detailed_arch}.
The cross attention between object queries and the feature maps~(either from point clouds or images) aggregates relevant context onto the object candidates, while the self attention between object queries reasons pairwise relations between different object candidates. The query positions are embedded into $d$-dimensional positional encoding with a Multilayer Perceptron~(MLP), {and element-wisely summed with the query features}. This enables the network to reason about both context and position jointly. 

The $N$ object queries containing rich instance information are then independently decoded into boxes and class labels by a feed-forward network~(FFN). 
 Following CenterPoint~\cite{Yin2020Centerbased3O}, our FFN predicts the center offset from the query position as $\delta x, \delta y$, bounding box height as $z$, size $l, w, h$ as $\log(l), \log(w), \log(h)$, yaw angle $\alpha$ as $sin(\alpha), cos(\alpha)$ and the velocity~(if available) as $v_x, v_y$. We also predict a per-class probability $\hat p \in [0,1]^K$ for $K$ semantic classes. Each attribute is computed by a separate two-layer $ 1\times 1$ convolution. 
 By decoding each object query into prediction in parallel, we get a set of predictions $\{\hat b_t, \hat p_t\}_{t}^N$ as output, where $\hat b_t$ is the predicted bounding box for the $i$-th query. 
Following \cite{misra20213detr}, we adopt the auxiliary decoding mechanism{,} which adds FFN and supervision after each decoder layer{. Hence}, 
we {can} 
have initial bounding box predictions from the first decoder layer. We leverage such initial predictions in the LiDAR-camera fusion module to constrain the cross attention, {as} 
explained in {the} next section. 




\subsection{LiDAR-Camera Fusion}

\noindent\textbf{Image Feature Fetching.} Although impressive improvement has been brought by point-level fusion methods~\cite{Vora2020PointPaintingSF, Wang2021PointAugmentingCA}, their fusion quality is largely limited by the sparsity of LiDAR points. When an object only contains a small number of LiDAR points, it can 
fetch {only} the same number of image features, wasting the rich semantic information of high-resolution images. To mitigate this issue, we do not fetch the multiview image features based on the hard association between LiDAR points and image pixels. Instead, we retain all the image features $F_C \in \mathbb{R}^{N_v \times H \times W \times d}$ as our memory bank, and use the cross-attention mechanism in {the} transformer decoder to perform 
feature fusion 
in a sparse-to-dense and adaptive manner, as shown in Fig.~\ref{fig:pipeline}. 

\begin{figure}[t]
	\vspace{-0.5cm}
	\setlength{\abovecaptionskip}{0.0cm}
	\setlength{\belowcaptionskip}{-0.45cm}
	\center
    \includegraphics[width=8cm]{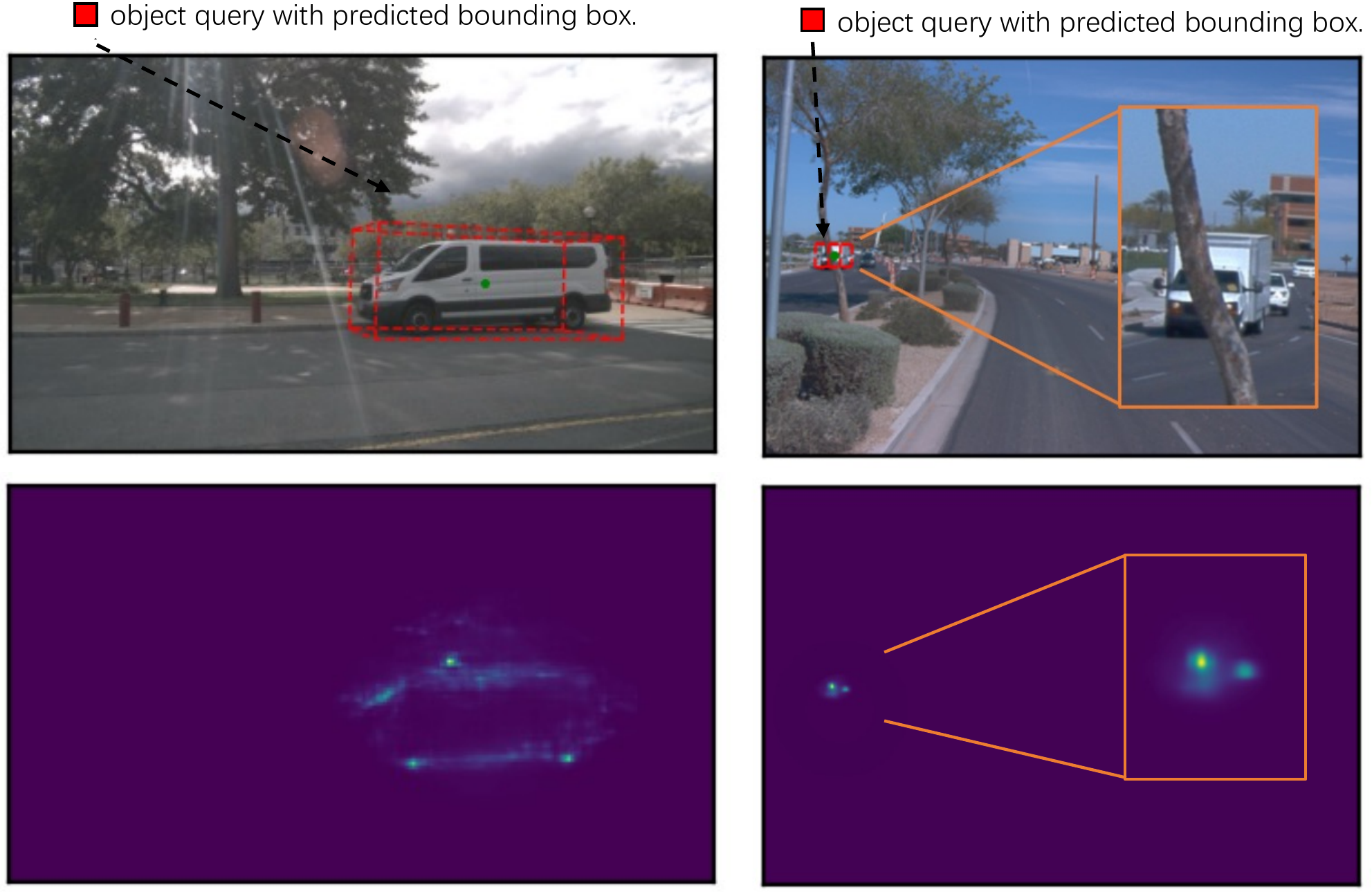}
    \caption{The first row shows the input images and the predictions of object queries projected on the images, and the second row shows the cross-attention maps. Our fusion strategy is able to dynamically choose relevant image pixels and is not limited by the number of LiDAR {points}. 
    The two images are picked from nuScenes and Waymo, respectively.}
    \label{fig:cross_attention}
\end{figure}

\noindent\textbf{SMCA for Image Feature Fusion.}
\label{subsubsec:img_fusion}
Multi-head attention is a popular mechanism to perform information exchange and build {a soft association} 
between two sets of inputs, and it has been widely used {for the feature matching task}
~\cite{Sarlin2020SuperGlueLF, Sun2021LoFTRDL}. To mitigate the sensitivity towards sensor calibration and inferior image features brought by the hard-association strategy, we leverage the cross-attention mechanism to build {the soft association} 
between LiDAR and images, enabling the network to adaptively determine where and what information should be taken from the images. 

Specifically, we first {identify} 
the specific image {in which the object queries are located} 
using previous predictions as well as the calibration matrices, and then perform cross attention between {the} object queries and the corresponding image feature map.
However, as the LiDAR features and image features are from completely different domains, the object queries might attend to visual regions unrelated to the bounding box to be predicted, leading to a long training time for the network to accurately identify the proper regions on images. Inspired by \cite{Gao2021FastCO}, we design a spatially
modulated cross attention~(SMCA) module{,} which weighs the cross attention by a 2D circular Gaussian mask around the projected 2D center of each query. The 2D Gaussian {weight} mask $\mathbf{M}$ is generated in a similar way as CenterNet~\cite{Zhou2019ObjectsAP},
	$M_{ij} = \exp(-\frac{(i-c_x)^2 + (j-c_y)^2}{\sigma r^2}),$
 where $(i,j)$ is the spatial indices of the weight mask $\mathbf{M}$, $(c_x, c_y)$ is the 2D center computed by projecting the query prediction 
 onto the image plane, $r$ is the radius of the minimum circumscribed circle
 of the projected corners of the 3D bounding box, and $\sigma$ is the hyper-parameter to modulate the bandwidth of the Gaussian distribution. 
Then this weight map is element-wisely multiplied with the cross-attention map among all the {attention} heads.
In this way, each object query only attends to the related region around the projected 2D box, so that the network can learn where to select image features based on the input LiDAR features better and faster. The visualization of the attention map is shown in Fig.~\ref{fig:cross_attention}. The network typically tends to focus on the foreground pixels close to the object center and ignore the irrelevant pixels, providing valuable semantic information for {object classification and bounding box regression.}
After SMCA, we use another FFN to produce the final bound box predictions using the object queries containing both LiDAR and image information.




\subsection{Label Assignment and Losses} 
\noindent Following DETR~\cite{misra20213detr}, we find the bipartite matching between {the} predictions and ground truth objects through {the} Hungarian algorithm~\cite{Kuhn1955TheHM}, where the matching cost is defined by a weighted sum of classification, regression, and IoU cost{:}
\begin{equation}
	C_{match} = \lambda_1 L_{cls}(p, \hat p) + \lambda_2 L_{reg}(b, \hat b) + \lambda_3 L_{iou}(b, \hat b),
	\label{eq:matching_cost}
\end{equation}
where $L_{cls}$ is the binary cross entropy loss, $L_{reg}$ is the L1 loss between {the} {predicted BEV centers and the ground-truth centers~(both normalized in $[0, 1]$),} 
and $L_{iou}$ is the IoU loss~\cite{Zhou2019IoULF} between {the} predicted boxes and ground-truth boxes. $\lambda_1, \lambda_2, \lambda_3$ are the coefficients of the individual cost terms. We provide sensitivity {analysis} 
of these terms in the supplementary Sec.~\ref{supp:label_assign}. Since the number of predictions is usually larger than that of GT boxes, the unmatched predictions are considered as negative samples.
Given all matched pairs, we compute {a} focal loss~\cite{Lin2017FocalLF} for the classification branch. The bounding box regression is supervised by {an} L1 loss for only positive pairs. For the heatmap prediction, we adopt a penalty-reduced focal loss following CenterPoint~\cite{Yin2020Centerbased3O}. The total loss is the weighted sum of losses for each component. We adopt the same label assignment strategy and loss formulation for both decoder layers.




\subsection{Image-Guided Query Initialization}
\label{subsubsec:img_query} 
{Since} our object queries are {currently} selected using only
LiDAR features, {it potentially leads to} 
sub-optimality in terms of the detection recall.
Empirically, our model already achieves high recall and shows superior performance over {the} 
baselines~(Sec.~\ref{sec:Exp}). Nevertheless, to further leverage the ability of high-resolution images 
{in detecting} {small} 
objects and make our algorithm more robust {against} 
sparse LiDAR point clouds, we propose an \secondmodule~strategy{,} which selects object queries leveraging both {the} LiDAR and camera information.
 
\begin{figure}[t]
	\vspace{-0.5cm}
	\setlength{\abovecaptionskip}{0.0cm}
	 \setlength{\belowcaptionskip}{-0.45cm}
	\center
    \includegraphics[width=8cm]{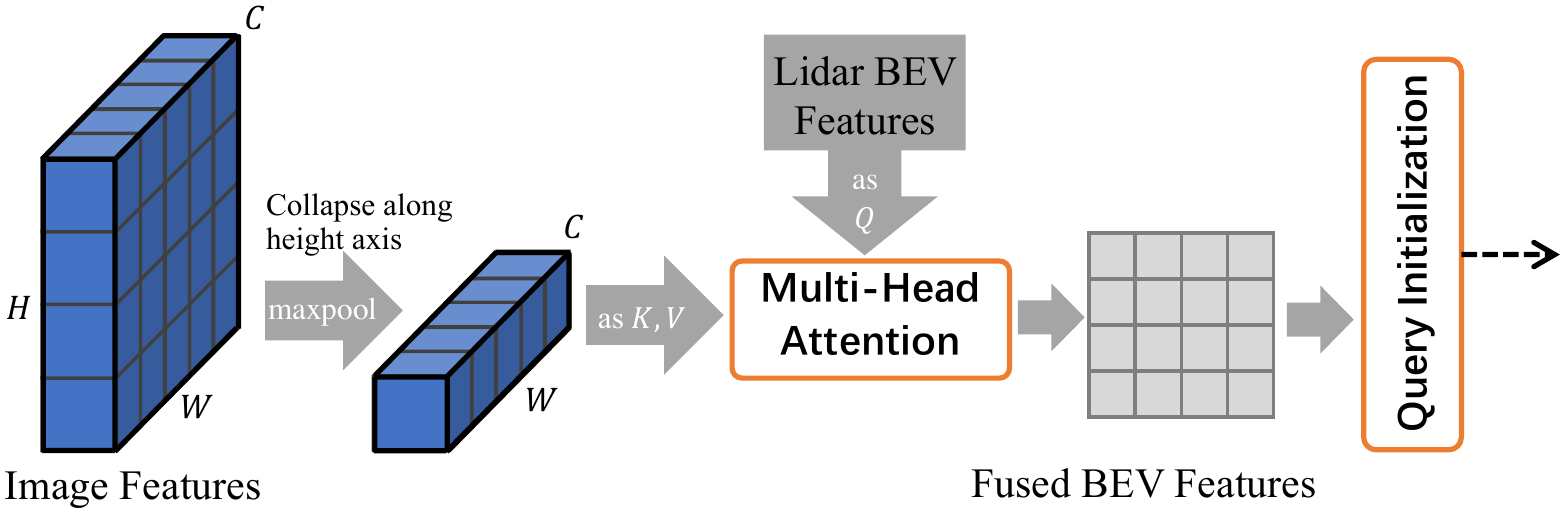}
    \caption{We first condense the image features along the vertical dimension, {and} then project the features onto {the} BEV plane using cross attention with the LiDAR BEV features. Each image is process{ed} by a separate multi-head attention layer{,} which captures the relation between image column and BEV locations.}
    \label{fig:img2bev}
\end{figure}

Specifically, we generate a LiDAR-camera BEV feature map $F_{LC}$ by projecting the image features $F_{C}$ onto {the} BEV plane through cross attention with LiDAR BEV features $F_{L}$.
Inspired by \cite{Roddick2020PredictingSM}, we use the multiview image features collapsed {a}long the height axis as 
{the key-value sequence of the attention mechanism,} 
as shown in Fig.~\ref{fig:img2bev}. The  {collapsing} operation is based on the observation that the relation between BEV locations and image columns can be established easily using camera geometry, and usually there is at most one object along each image column. Therefore, collapsing {along} the height axis {can} 
significantly reduce the computation without losing critical information. Although some fine-grained image features might be lost during this process, it already 
{meets} our need 
{as only a hint on potential object positions is required.} 
Afterward, similar {to} 
Sec. \ref{subsubsec: query_initialize}, we use $F_{LC}$ to predict the heatmap{, which is} 
averaged with the LiDAR-only heatmap $\hat S$ as the final heatmap $\hat S_{LC}$. Using $\hat S_{LC}$ to select and initialize the object queries, our model is able to detect objects that are difficult to detect in {LiDAR point clouds.}

Note that proposing a novel method to project the image features onto the BEV plane is beyond the scope of this paper{. We} 
believe that our method could benefit from more research progress~\cite{Roddick2019OrthographicFT, Roddick2020PredictingSM, Philion2020LiftSS} 
{in} this direction.

\xy{
}

 \section{Implementation Details}

\begin{table*}[ht]
\vspace{-0.6cm}
\setlength{\abovecaptionskip}{0.0cm}
\setlength{\belowcaptionskip}{-0.45cm}
	\centering
	\resizebox{1.0\textwidth}{!}{
	\begin{tabular}{l|c|c|cc|cccccccccc}
		\Xhline{2\arrayrulewidth}
         {Method} & {Modality} &{Voxel Size~(m)}& {mAP} & {NDS} & {Car} & {Truck} & {C.V.} & {Bus} & {Trailer} & {Barrier} & {Motor.} & {Bike} & {Ped.} & {T.C.} \\
        \hline
         \textbf{PointPillar\cite{Lang2019PointPillarsFE}} & L & $(0.2, 0.2, 8)$ & 40.1 & 55.0 & 76.0 & 31.0 & 11.3 & 32.1 & 36.6 & 56.4 & 34.2 & 14.0 & 64.0 & 45.6 \\
          \textbf{CBGS\cite{Zhu2019ClassbalancedGA}} & L & $(0.1, 0.1, 0.2)$ & 52.8 & 63.3 & 81.1 & 48.5 & 10.5 & 54.9 & 42.9 & 65.7 & 51.5 & 22.3 & 80.1 & 70.9 \\
          \textbf{CenterPoint\cite{Yin2020Centerbased3O}} & L & $(0.075, 0.075, 0.2)$ & 60.3 & 67.3 & 85.2 & 53.5 & 20.0 & 63.6 & 56.0 & 71.1 & 59.5 & 30.7 & 84.6 & 78.4 \\
		\hdashline
          \textbf{PointPainting\cite{Vora2020PointPaintingSF}} & LC & $(0.2, 0.2, 8)$  & 46.4 & 58.1 & 77.9 & 35.8 & 15.8 & 36.2 & 37.3 & 60.2 & 41.5 & 24.1 & 73.3 & 62.4 \\
          \textbf{3D-CVF\cite{Yoo20203DCVFGJ}} & LC & $(0.05, 0.05, 0.2)$  & 52.7 & 62.3 & 83.0 & 45.0 & 15.9 & 48.8 & 49.6 & 65.9 & 51.2 & 30.4 & 74.2 & 62.9 \\
          \textbf{PointAugmenting\cite{Wang2021PointAugmentingCA}} & LC & $(0.075, 0.075, 0.2)$  & 66.8 & 71.0 & \red{87.5} & 57.3 & 28.0 & 65.2 & 60.7 & 72.6 & 74.3 & 50.9 & 87.9 & 83.6 \\
          \textbf{MVP\cite{Yin2021MVP}} & LC & $(0.075, 0.075, 0.2)$ & 66.4 & 70.5  & 86.8 & 58.5 & 26.1 & 67.4 & 57.3 & 74.8 & 70.0 & 49.3 & \red{89.1} & 85.0 \\
          \textbf{FusionPainting\cite{Xu2021FusionPaintingMF}} & LC & $(0.075, 0.075, 0.2)$  & 68.1 & 71.6 & 87.1 & \red{60.8} & 30.0 & \red{68.5} & \red{61.7} & 71.8 & \red{74.7} & \red{53.5} & 88.3 & 85.0 \\
		\hdashline
         \textbf{\Name-L} & L & $(0.075, 0.075, 0.2)$ & \blue{65.5} & \blue{70.2} & \blue{86.2} & \blue{56.7} & \blue{28.2} & \blue{66.3} & \blue{58.8} & \blue{78.2} & \blue{68.3} & \blue{44.2} & \blue{86.1} & \blue{82.0} \\
         \textbf{\Name} & LC & $(0.075, 0.075, 0.2)$ & \red{68.9} & \red{71.7} & 87.1 & 60.0 & \red{33.1} & 68.3 & 60.8 & \red{78.1} & 73.6 & 52.9 & {88.4} & \red{86.7} \\
         
         \Xhline{2\arrayrulewidth}
	\end{tabular}
	}
	\caption[STH]{Comparison with SOTA methods on the nuScenes test set. ‘C.V.’, ‘Ped.’, and ‘T.C.’ are  short for construction vehicle, pedestrian, and traffic cone, respectively. ‘L’ {and} 
	‘C’ represent LiDAR and Camera, respectively. The best results are {in boldface} 
	(Best LiDAR-only results are marked \blue{blue} and best LC results are marked \red{red}). For FusionPainting~\cite{Xu2021FusionPaintingMF}, we report the results on the nuScenes website{, which are} 
	better than what they reported in {their} paper. 
 	Note that CenterPoint~\cite{Yin2020Centerbased3O} and PointAugmenting~\cite{Wang2021PointAugmentingCA} utilize double-flip testing while we {do} not use any test time augmentation. 
	{Please find detailed} 
	results
	here.\footnotemark }
	\label{tab:nuscene_test}
\end{table*}

\label{sec:imp}
\noindent \textbf{Training.} We implement our network in PyTorch~\cite{paszke2017automatic} using {the} open-sourced MMDetection3D~\cite{mmdet3d2020}. For nuScenes, we use the DLA34\cite{Yu2018DeepLA} of the pretrained CenterNet as our 2D backbone and keep {its weights} 
frozen during training, following \cite{Wang2021PointAugmentingCA}. We set the image size to $448 \times 800$, which performs {comparably} 
with full resolution{~($896\times 1600$)}.
VoxelNet~\cite{Zhou2018VoxelNetEL, Yan2018SECONDSE} is chosen as our 3D backbone. 
Our training consists of two stages: 1) We first train the 3D backbone with the first decoder layer and FFN for 20 epochs, which only need{s} the LiDAR point clouds as input and produce{s} the initial 3D bounding box predictions. We adopt the same data augmentation and training schedules as prior LiDAR-only works~\cite{Yin2020Centerbased3O, Zhu2019ClassbalancedGA}. Note that we also find the copy-and-paste augmentation strategy~\cite{Yan2018SECONDSE} benefits the convergence but could disturb the real data distribution, so we disable this augmentation for {the} last 5 epochs following~\cite{Wang2021PointAugmentingCA} {(they called {a} fade strategy)}. 2) We then train the LiDAR-camera fusion and the \secondmodule~module for another 6 epochs. We find that this {two-step} 
training scheme performs better {than joint training}, since we {can} 
adopt more flexible augmentations for the first training stage. See supplementary Sec.~\ref{supp:imp_detail}
for {the} detailed hyper-parameters and settings on Waymo.


\noindent\textbf{Testing.} {During inference, the final score is computed as the geometric average of the heatmap score $\hat S_{ij}$ and the classification score $\hat p_t$. We use all the outputs as our final predictions without Non-maximum Suppression~(NMS) (see 
the effect of NMS 
in supplementary Sec.~\ref{supp:nms})}. It is noteworthy that previous point-level fusion methods such as PointAugmenting~\cite{Wang2021PointAugmentingCA} {rely} 
on two different models for camera FOV and LiDAR-only region{s} if the cameras are not 360-degree cameras, 
because only points in the camera FOV could fetch the corresponding image features. In contrast, we use a single model to deal with both camera FOV and LiDAR-only region{s, since} 
object queries locate{d outside} 
camera FOV will directly ignore the fusion stage and the initial predictions from the first decoder layer will be a safeguard. 
\section{Experiments}
	\label{sec:Exp}
In this section, we first make comparisons with {the} state-of-the-art methods on nuScenes and Waymo.
Then we conduct extensive ablation studies to demonstrate the importance of each {key} component {of \Name}. Moreover, we design two experiments to show the robustness of our \Name~{against} inferior image conditions. 
{Besides \Name, we also include a model variant{, which is based on} 
the first training stage, {i.e., producing} 
the initial bounding box predictions using only point clouds. We denote it as \Name-L and believe that it can serve as a strong baseline for LiDAR-only detection.} {We provide the qualitative results in supplementary Sec.~\ref{supp:visualization}}.

\noindent\textbf{nuScenes Dataset.}
\footnotetext{\noindent\url{https://www.nuscenes.org/object-detection}}
The nuScenes dataset is a large-scale autonomous-driving dataset for 3D detection and tracking, consisting of 700, 150, {and} 150 scenes for training, validation, and testing, respectively. 
Each frame contains one point cloud and six calibrated images covering the 360{-}degree {horizontal} FOV. For 3D detection, the main metrics are mean Average Precision~(mAP)~\cite{Everingham2009ThePV} and nuScenes detection score~(NDS). The mAP is defined by the BEV center distance instead of the 3D IoU, and the final mAP is computed by averaging over distance thresholds of $0.5m, 1m, 2m, 4m$ across ten classes. NDS is a consolidated metric of mAP and other attribute metrics, including translation, scale, orientation, velocity, and other box attributes. 
Following CenterPoint~\cite{Yin2020Centerbased3O}, we set the voxel size to $(0.075m, 0.075m, 0.2m)$.

\noindent\textbf{Waymo Open Dataset.} {This dataset}
consists of 798 scenes for training and 202 scenes for validation. The official metrics are mAP and mAPH~(mAP weighted by heading accuracy). The mAP and mAPH are defined based on the 3D IoU threshold of 0.7 for vehicles and 0.5 for pedestrians and cyclist{s}. These metrics are further broken down into two difficulty levels: LEVEL1 for boxes with more than five LiDAR points and LEVEL2 for boxes with at least one LiDAR point. Unlike the 360-degree cameras in nuScenes, the cameras in Waymo only cover around 250 degree{s} horizontally.
The voxel size is set to $(0.1m, 0.1m, 0.15m)$.

\subsection{Main Results}
\label{subsec:main_res}

\noindent\textbf{nuScenes Results.} We submitted our detection results to the nuScenes evaluation server. 
{Without any test time augmentation or model ensemble, our \Name~outperforms all competing non-ensembled methods on the nuScenes leaderboard at the time of submission.} 
As shown in Table~\ref{tab:nuscene_test}, our \Name-L already outperforms the 
state-of-the-art LiDAR-only methods by a significant margin~(+5.2\% mAP, +2.9\% NDS) and even surpasses some multi-modality methods. We ascribe this performance gain to the relation modeling power of {the} transformer decoder as well as the proposed query initialization strategies, which {are} 
ablated in Sec.~\ref{sub:ablation}.
Once enabling the proposed fusion components, our \Name~receives remarkable performance boost~(+3.4\% mAP, +1.5\% NDS) and {outperforms} 
all the previous methods, including {FusionPainting} \cite{Xu2021FusionPaintingMF}{,} 
which uses {extra data}
{to train their} segmentation {sub-}networks.
Moreover, thanks to our soft-association mechanism, \Name~is robust to inferior image conditions including degenerated image quality and sensor misalignment, as shown in the next section.

\begin{table}[t]
\vspace{-0.4cm}
\setlength{\abovecaptionskip}{0.0cm}
\setlength{\belowcaptionskip}{-0.35cm}
    \centering
    \resizebox{0.45\textwidth}{!}{
        \begin{tabular}{l|cccc}
         \Xhline{2\arrayrulewidth}
        \textbf{} & \textbf{Vehicle} & \textbf{Pedestrian} & \textbf{Cyclist} & \textbf{Overall} \\
        \hline
        \textbf{PointPillar\cite{Vora2020PointPaintingSF}} & 62.5 & 50.2 & 59.9 & 57.6 \\
        \textbf{PVRCNN\cite{Shi2020PVRCNNPF}} & 64.8 & 46.7 & - & -  \\
        \textbf{LiDAR-RCNN\cite{Li2021LiDARRA}} & 64.2 & 51.7 & 64.4 & 60.1 \\
        \textbf{CenterPoint\cite{Yin2020Centerbased3O}} & 66.1 & 62.4 & 67.6 & 65.3 \\
        \textbf{PointAugmenting\cite{Wang2021PointAugmentingCA}} & 62.2 & 64.6 & 73.3 & 66.7 \\
        \hdashline
        \textbf{\Name-L} & 65.1 & 63.7 & 65.9 & 64.9 \\
        \textbf{\Name} & 65.1 & 64.0 & 67.4 & 65.5 \\
        \Xhline{2\arrayrulewidth}
        \end{tabular}
    }
    \caption[STH]{LEVEL\_2 mAPH on Waymo validation set. For CenterPoint, we report the performance of single-frame one-stage model trained in 36 epochs. 
    }
    \label{tab:waymo_val}
\end{table}

\noindent\textbf{Waymo Results.} We report the performance of our model over all three classes on Waymo validation set in Table~\ref{tab:waymo_val}. 
{Our fusion strategy {improves the mAPH of pedestrian and cyclist classes by 0.3 and 1.5x,} 
respectively.}
 We suspect two reasons for the relatively small improvement brought by the image components. 
 First, the semantic information of images might have less impact on the coarse-grained categorization of Waymo.
Second, the initial bounding boxes from {the} first decoder layer are already with accurate locations since the point clouds in Waymo are denser than {those} 
in nuScenes {(see more discussions in supplementary Sec.~\ref{supp:waymo_dis}).}
Note that CenterPoint
achieves a better performance with a multi-frame input and a second-stage refinement module. Such components are orthogonal {to} 
our method and we leave a more powerful \Name~for Waymo as the future work. {PointAugmenting achieves better performance than ours but relies on CenterPoint to get the predictions outside camera FOV for a full-region detection, making their system less flexible.}
%


\noindent\textbf{Extend to Tracking.} To further demonstrate the generalization capability, we evaluate our model in {a} 
3D multi-object tracking~(MOT) task by performing tracking-by-detection with the same tracking algorithms adopted by CenterPoint. We refer readers to the original paper \cite{Yin2020Centerbased3O} for details. As shown in Table~\ref{tab:tracking}, our model significantly outperforms CenterPoint and sets 
{the new state-of-the-art results}
on the leaderboard of nuScenes tracking.

\begin{table}[t]
\setlength{\abovecaptionskip}{0.0cm}
\setlength{\belowcaptionskip}{-0.35cm}
    \centering
    \resizebox{0.45\textwidth}{!}{
        \begin{tabular}{l|ccccc}
         \Xhline{2\arrayrulewidth}
        \textbf{} & \textbf{AMOTA$\uparrow$} & \textbf{TP$\uparrow$} & \textbf{FP$\downarrow$} & \textbf{FN$\downarrow$} & \textbf{IDS$\downarrow$}\\
        \hline
        \textbf{CenterPoint\cite{Yin2020Centerbased3O}} & 63.8 & 95877 & 18612 & 22928 & 760  \\
        \textbf{EagerMOT\cite{Kim2021EagerMOT3M}} & 67.7 & 93484 & 17705 & 24925 & 1156  \\
        \textbf{AlphaTrack\cite{Zeng2021AlphaTrack}} & 69.3 & 95851 & 18421 & 22996 & \textbf{718}  \\
        \hdashline
        \textbf{\Name-L} & 68.6 & 95235 & 17851 & 23437 & 893 \\
        \textbf{\Name} & \textbf{71.8} &  \textbf{96775} & \textbf{16232} & \textbf{21846} & 944 \\

        \Xhline{2\arrayrulewidth}
        \end{tabular}
    }
    \caption[STH]{Comparison of the tracking results on nuScenes test set. Please find detailed results here.\footnotemark
    }
    \label{tab:tracking}
\end{table}
\footnotetext{\url{https://www.nuscenes.org/tracking}}




\subsection{Robustness against Inferior Image Conditions} 
\label{subsec:robust}
We design three experiments to demonstrate the robustness of our proposed fusion module. Since the nuScenes test set only allows at most three submissions, all the experiments are conducted on the validation set. For fast iteration, we reduce the first stage training to 12 epochs and remove the fade strategy. All the other parameters are the same as the main experiments. {To avoid overstatement,} {we additionally build two baseline LiDAR-camera detectors by equipping our \Name-L with two representative fusion methods on nuScenes: fusing LiDAR and image features by point-wise concatenation~(denoted as \textbf{CC}) and the fusion strategy of PointAugmenting~(denoted as \textbf{PA}). 
}

\noindent\textbf{Nighttime.} We first split the validation set into daytime and nighttime based on scene descriptions provided by nuScenes and show the performance gain under different situations in Table~\ref{tab:abl_daynight}. Our method brings {a} much larger performance gain during nighttime,  where the worse lighting negatively affects the hard-association based fusion strategies {\textbf{CC} and \textbf{PA}.} 

\begin{table}[t]
\vspace{-0.4cm}
\setlength{\abovecaptionskip}{0.0cm}
\setlength{\belowcaptionskip}{-0.35cm}
	\centering
	\resizebox{0.35\textwidth}{!}{
	\begin{tabular}{l|cc}
		\Xhline{2\arrayrulewidth}
         & {Nighttime} & {Daytime} \\
        \hline
        \textbf{\Name-L} & 49.2 & 60.3 \\
        \hdashline
       	 \textbf{CC}     & 49.4 (+0.2) & 63.4 (+3.1) \\
       	\textbf{PA}      & 51.0 (+1.8) & 64.3 (+4.0) \\
       	\textbf{\Name}   & 55.2 (+6.0) & 65.7 (+5.4) \\
         \Xhline{2\arrayrulewidth}
	\end{tabular}
	}
	\caption{mAP breakdown over daytime and nighttime. We exclude categories that do no have any labeled samples.}
	\label{tab:abl_daynight}
\end{table}
%

\noindent\textbf{Degenerated Image Quality.} In Table~\ref{tab:abl_drop_img}, we randomly drop several images 
{for each frame} 
by setting the image features of {such images} 
to zero during inference. {Since} both \textbf{CC} and \textbf{PA} fuse LiDAR and image features in a tightly-coupled way, {their} 
performance drops significantly when some images are not available during inference. In contrast, our \Name~is able to maintain a high mAP under all cases. When all the six images are not available, \textbf{CC} and \textbf{PA} {suffer} 
from $23.8\%$ and $17.2\%$ mAP degradation, respectively, while \Name~still keeps the mAP at a competitive level of $61.7\%$. {This advantage comes from the sequential design and the attentive fusion strategy{, which} 
first generate{s} initial predictions based on LiDAR data and then only gathers useful information from image features adaptively. Moreover, we could {even} directly {disable} 
the fusion module if the camera malfunctioning is known, such that the whole system could still work seamlessly in a LiDAR-only mode.} 

\begin{table}[h]
\setlength{\abovecaptionskip}{0.0cm}
\setlength{\belowcaptionskip}{-0.25cm}
    \centering
    \resizebox{0.48\textwidth}{!}{
        \begin{tabular}{l|cccc}
         \Xhline{2\arrayrulewidth}
        \textbf{\# Dropped Images} & \textbf{0} & \textbf{1} & \textbf{3} & \textbf{6}\\
        \hline
        \textbf{CC} & 63.3 & 59.8~(-3.5) & 50.9~(-12.4) & 39.5~(-23.8) \\
        \textbf{PA} & 64.2 & 61.6~(-2.6) & 55.4~(-8.8) & 47.0~(-17.2) \\
        \textbf{\Name} & 65.6 & 65.1~(-0.5) & 63.9~(-1.7) & 61.7~(-3.9)  \\

        \Xhline{2\arrayrulewidth}
        \end{tabular}
    }
    \caption{mAP under different numbers of dropped images. {The number in each bracket} 
    is the mAP drop from the standard input. }
    \label{tab:abl_drop_img}
\end{table}

\begin{figure}[t]
\vspace{-0.35cm}
\setlength{\abovecaptionskip}{0.0cm}
\setlength{\belowcaptionskip}{-0.25cm}
\centering
    \includegraphics[width=6.5cm]{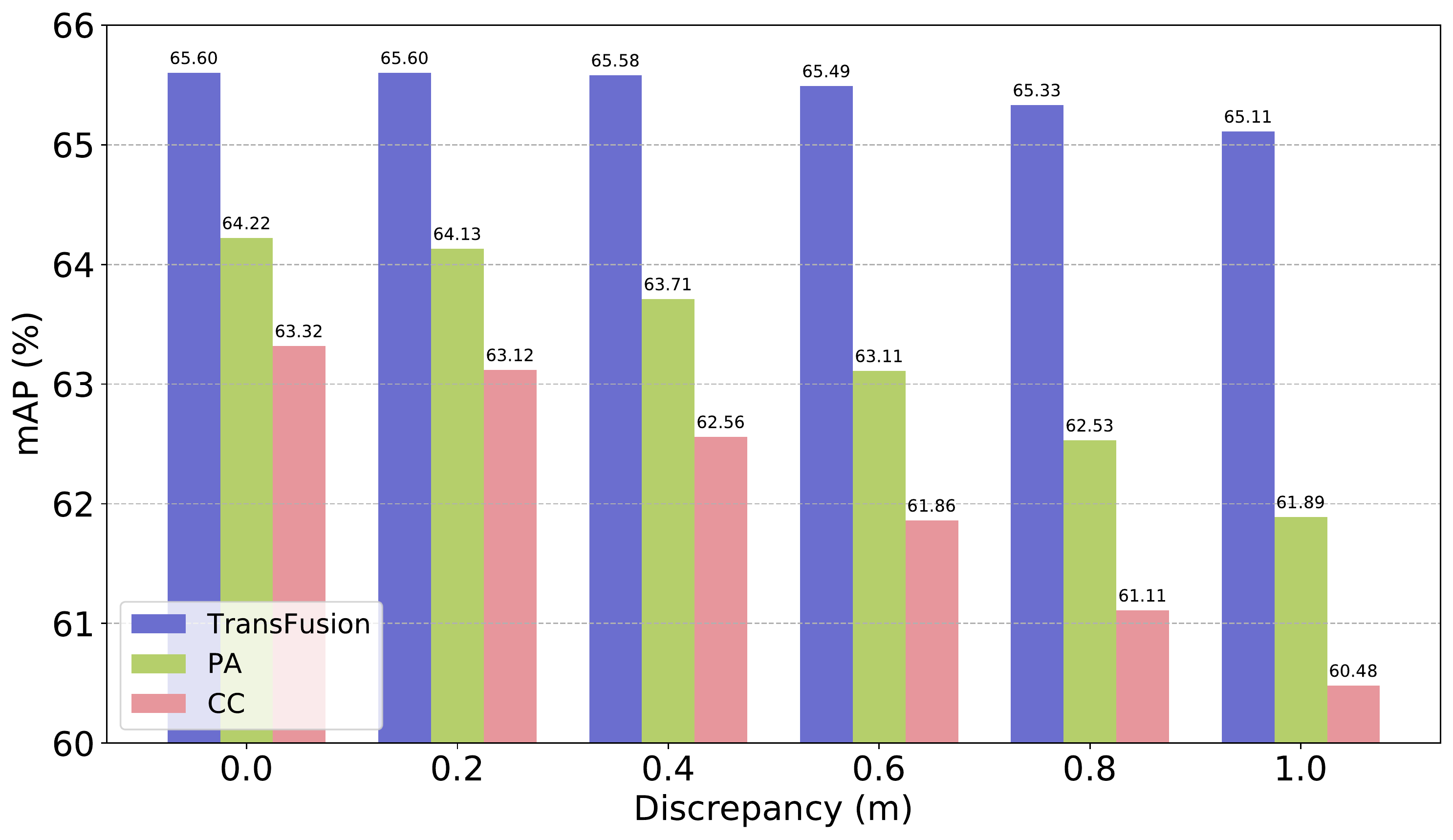}
    \caption{mAP under sensor misalignment cases. The X axis refers to the translational discrepancy between two sensors.
      }
    \label{fig:abl_calibration}
\end{figure}

\noindent\textbf{Sensor Misalignment.} 
We evaluate different fusion methods under a setting where LiDAR and images are not well-calibrated following RoarNet~\cite{Shin2019RoarNetAR}. Specifically, we randomly add a translation offset to the transformation matrix from camera to LiDAR sensor. As shown in Fig.~\ref{fig:abl_calibration}, \Name~achieves better robustness against the calibration error compared with other fusion methods. When two sensors are misaligned
by $1m$, the mAP of our model only drops by 0.49\%, while {the} mAP of \textbf{PA} and \textbf{CC} degrades by 2.33\% and 2.85\%, respectively. In our method, the calibration matrix is only used for projecting the object queries onto images, 
{and the fusion module} is not strict with the projected locations {since} the attention mechanism could adaptively find the relevant image features around based on the context information. The insensitivity towards sensor calibration also enables the possibility to pipelining 
the 2D and 3D backbones such that the LiDAR features are fused with the features from the previous images~\cite{Vora2020PointPaintingSF}.

\subsection{Ablation Studies}
\label{sub:ablation}
We conduct ablation studies on the nuScenes {validation set} to study the effectiveness of the proposed components. 

\begin{table}[h]
\setlength{\abovecaptionskip}{0.0cm}
\setlength{\belowcaptionskip}{-0.35cm}
	\centering
	\resizebox{0.4\textwidth}{!}{
	\begin{tabular}{l|cccc|cc}
		\Xhline{2\arrayrulewidth}
         & {C.A.} & {I.D.} & {\#Layers} & {\#Epochs} & {mAP} & {NDS} \\
        \hline
        a)  & \checkmark & \checkmark & 1 & 12 & 60.0 & 66.8 \\
        b) &   & \checkmark & 1 & 12 & 54.3 & 63.9 \\
        c)  & \checkmark & \checkmark & 3 & 12 & 59.9 & 67.1 \\
        \hdashline
        d) & \checkmark  &  & 1 & 12 & 24.0 & 33.8 \\
        e) & \checkmark  &  & 3 & 12 & 28.3 & 43.4 \\
        f) & \checkmark  &  & 3 & 36 & 46.9 & 57.8 \\
         \Xhline{2\arrayrulewidth}
	\end{tabular}
	}
	\caption{Ablation of the query initialization module. C.A.: category-aware; I.D.: \ourqueryinit. 
	}
	\label{tab:abl_lidar}
\end{table}

\noindent\textbf{Query Initialization.} In Table~\ref{tab:abl_lidar}, we study how the query initialization strategy affect{s} the performance of the initial bounding box prediction. \textbf{a)} the first row is \Name-L. \textbf{b)} when the category-embedding is removed, NDS drops to 63.9\%. \textbf{d)-f)} shows the performance of the models trained without the \ourqueryinit~strategy. Specifically, we make the query positions as a set of learnable parameters~($N\times 2$) to capture {the} statistics of potential object locations in the dataset.
The model under this setting only achieves 33.8\% NDS. Increasing {the} number of decoder layers
{or the number of} 
training epochs 
boosts the performance, but {\Name-L still outperforms the model in (f)} 
by 9.0\% NDS.
\textbf{a), c)}:
{In contrast, with the proposed query initialization strategy,  our \Name-L does not require more decoder layers.}

\noindent\textbf{Fusion Components.}
{To study how the image information benefits the detection results, we ablate the proposed fusion components by removing the feature fusion module~(denoted as \textbf{w/o Fusion}) and the \secondmodule~(denoted as \textbf{w/o Guide}).}
As shown in Table~\ref{tab:abl_fusion}, the image feature fusion and \secondmodule~bring 4.8\% and 1.6\% mAP gain, respectively. The former provides more distinctive instance features{, which are} 
particularly critical for classification on nuScenes, where some categories are challenging to distinguish, such as trailer and construction vehicle. {The} 
latter affects less{, since} 
\Name-L already has enough recall.  {We} 
believe the latter 
will be more useful when 
point clouds are sparser. 
Compared with other fusion method{s}, our fusion strategy brings a larger performance gain with {a modestly increasing number} {of parameters and latency}. 
To better understand where the improvements are from, we show the mAP breakdown on different subsets based on the range in Table~\ref{tab:abl_range}. Our fusion method gives larger performance boost for distant regions where 
3D objects are difficult to 
{detect or classify in}
LiDAR modality.

\begin{table}[t]
\vspace{-0.35cm}
\setlength{\abovecaptionskip}{0.0cm}
\setlength{\belowcaptionskip}{-0.45cm}
	\centering
	\resizebox{0.45\textwidth}{!}{
	\begin{tabular}{l|cc|cc}
		\Xhline{2\arrayrulewidth}
         & {mAP} & {NDS} & {Params~(M)} & {Latency~(ms)}  \\
        \hline
        \textbf{CenterPoint} & 57.4 & 65.2 & 8.54 & 117.2 \\
        \textbf{\Name-L} & 60.0 & 66.8 & 7.96 & 114.9  \\
         \hdashline
        \textbf{CC} & 63.3 & 67.6 & 8.01 + 18.34 &  212.3 \\
       	\textbf{PA} & 64.2 & 68.7 & 13.9 + 18.34&  288.2  \\
        \hdashline
       	\textbf{w/o Fusion} & 61.6 & 67.4 & 9.08 + 18.34 & 215.0 \\
       	\textbf{w/o Guide} & 64.8 & 69.3 & 8.35 + 18.34 & 236.9 \\
       	\textbf{\Name} & 65.6 & 69.7 &  9.47 + 18.34 & 265.9 \\
         \Xhline{2\arrayrulewidth}
	\end{tabular}
	}
	\caption{Ablation of the proposed fusion components. {18.34 represents the parameter size of the 2D backbone.}
	The latency is measured on an Intel Core i7 CPU and a Titan V100 GPU. For CenterPoint, we use re-implementations in MMDetection3D.
	}
	\label{tab:abl_fusion}
\end{table}

\begin{table}[ht]
\setlength{\abovecaptionskip}{0.0cm}
\setlength{\belowcaptionskip}{-0.45cm}
	\centering
	\resizebox{0.4\textwidth}{!}{
	\begin{tabular}{l|cccc}
		\Xhline{2\arrayrulewidth}
         & {$<$15m} & {15-30m} & {$>$30m} \\
        \hline
        \textbf{\Name-L} & 70.4 & 59.5 & 35.3 \\
       	\textbf{\Name}   & 75.5 (+5.1) & 66.9 (+7.4) & 43.7 
       	(+8.4) \\
         \Xhline{2\arrayrulewidth}
	\end{tabular}
	}
	\caption{mAP breakdown over BEV distance between object center and ego vehicle in meters.}
	\label{tab:abl_range}
\end{table}




\section{Conclusion}
{We have designed {an effective and robust} 
transformer-based 
LiDAR-camera 3D detection framework with a soft-association mechanism to adaptively determine where and what information should be taken from images.}
Our \Name~sets the new state-of-the-art results on the nuScenes detection and tracking leaderboard{s}, and shows competitive results on Waymo detection benchmark. The extensive ablative experiments demonstrate the robustness {of our method against} 
inferior image conditions.
{We hope that our work will inspire further investigation of LiDAR-camera fusion for driving-scene perception, and the application of {a} soft-association based fusion strategy to other tasks, such as 3D segmentation.}


\noindent\textbf{Acknowledgements.} This work is supported by Hong Kong RGC (GRF 16206819, 16203518, T22-603/15N), Guangzhou Okay Information Technology with the project GZETDZ18EG05, and City University of Hong Kong (No. 7005729).

\newpage
{\small
\bibliographystyle{ieee_fullname}
\bibliography{egbib}
}

\setcounter{page}{1}
\clearpage
{ 
\section*{Supplementary Material}
  \renewcommand\thesection{\Alph{section}}
\setcounter{section}{0}

	The supplementary document is organized as follows:
	\begin{itemize}[itemsep=-1mm]
	\item Sec.~\ref{supp:detailed_arch} depicts the detailed network architectures of our transformer decoder layers.
	\item Sec.~\ref{supp:imp_detail} provides the implementation details of \Name~and the training settings on nuScenes and Waymo. 
	\item Sec.~\ref{supp:label_assign} {reports our} 
	sensitivity analysis of the matching cost during label assignment.
	\item Sec.~\ref{supp:nms} presents the effect of NMS on \Name~and CenterPoint. 
	\item Sec.~\ref{supp:pillar} provides the results of using PointPillars as our 3D backbone.
	\item Sec.~\ref{supp:2dnet} discusses the effect of 2D backbone in \Name.
	\item Sec.~\ref{supp:num_query} shows the results with different number of object queries.
	\item Sec.~\ref{supp:waymo_dis} discusses the performance gain of image information on Waymo.
	\item Sec.~\ref{supp:visualization} provides visualization results on the nuScenes and Waymo datasets.
	\end{itemize}

\section{Network {A}rchitecture{s}}
\label{supp:detailed_arch}
The detailed architectures of {the respective} 
transformer decoder layers for initial bounding box prediction and LiDAR-camera fusion are 
 shown in Fig.~\ref{fig:detailed_arch}. {Following~\cite{Liu2021GroupFree3O},} we adopt the common practice of transformer except that we use {the} learned positional encoding instead of {the} fixed sine positional encoding~\cite{Vaswani2017AttentionIA}. 
For the \secondmodule~module, we use the LiDAR BEV features as query sequence and collapsed image feature{s} as key-value sequence, and only perform cross attention to save the computation cost. Our model can 
benefit from the efficient attention mechanism{s} in recent works such as \cite{Zhu2021DeformableDD}.

\begin{figure}[h]
\hspace{-0.25cm}
    \includegraphics[width=9cm]{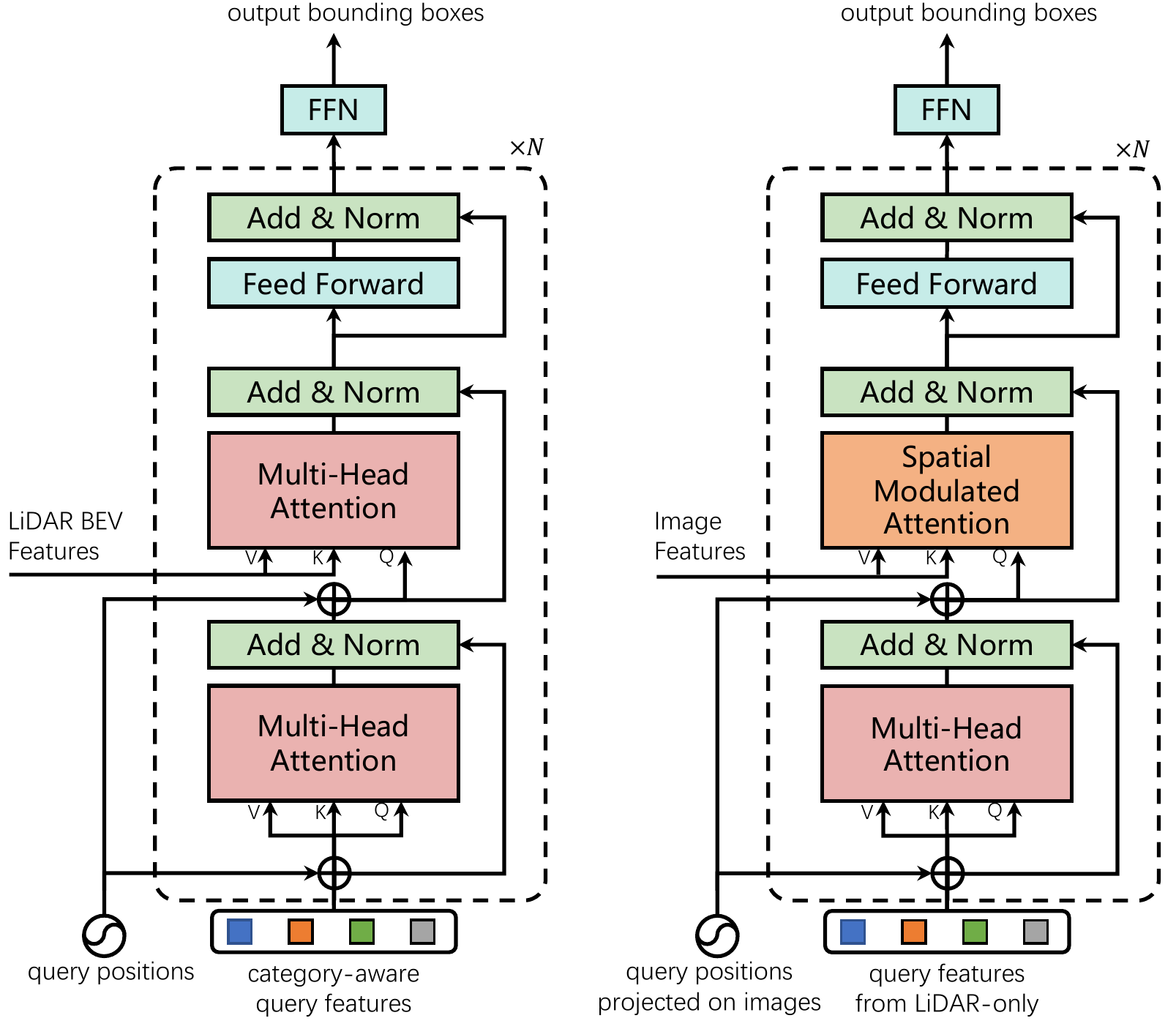}
    \caption{
    Left: Architecture of the transformer decoder layer for initial bounding box prediction. Right: Architecture of the transformer decoder layer for image fusion.
    }
    \label{fig:detailed_arch}
\end{figure}


\section{Implementation Details}
\label{supp:imp_detail}
Our implementation is based on the open-sourced codebase {MMDetection3D}
~\cite{mmdet3d2020}, which provide{s} many popular 3D detection methods, including PointPillar, VoxelNet, and CenterPoint. For {our} 3D backbone, we set {its} 
hyper-parameters according to CenterPoint-Voxel's official implementation. 
For the transformer-decoder-based detection head, the hidden dimension $d$ is set to 256 and dropout is set to $0.1$. {We use $N=200$ and $300$ queries for nuScenes and Waymo since the max number{s} of objects in one frame {are} 
142 and 185, respectively}. Since our object queries are non-parametric, we are able to modify the number of queries during inference. We provide the ablations on the number of object queries in Sec.~\ref{supp:num_query}. When selecting object queries from the heatmap, we pick local maximum pixels whose value{s are} 
greater than or equal to {their} 
8-connected neighbors. To avoid mistakenly suppress 
nearby instances for small objects, we {do} 
not check the local maximum for pedestrian and traffic cone on nuScenes {and for pedestrian and cyclist on Waymo}. Following PointAugmenting~\cite{Wang2021PointAugmentingCA}, we adopt DLA34 of CenterNet pre-trained on monocular 3D detection task as our 2D backbone for nuScenes. {Since there is no public available 2D backbone pre-trained on Waymo dataset, we train a Faster-RCNN~\cite{Ren2015FasterRT} using the 2D labels provided by Waymo and use its ResNet50 and FPN as our 2D backbone}. We freeze the weight of image backbone during training and 
set the image resolution to half of the full resolution for both nuScenes and Waymo to speed up the training process.

\noindent\textbf{nuScenes.} Following the common practice, we transform previous ten LiDAR sweeps into 
{the} current frame to produce a denser point cloud input for both training and inference. The detection range is set to $[-51.2m, 51.2m]$ for X and Y {axes}, 
and $[-5m, 3m]$ for Z axis. The maximum number{s} of non-empty voxel{s for training and inference are} 
set to 120,000 and 160,000, 
respectively. In terms of the data augmentation strategy, we adopt random flipping along both X and Y {axes}, 
global scaling with {a} random factor from $[0.9, 1.1]$, global rotation between $[-\pi/8, \pi/8]$, as well as the 
copy-and-paste augmentation~\cite{Yan2018SECONDSE}.  We follow CBGS~\cite{Zhu2019ClassbalancedGA} to perform class-balanced sampling and optimize the network using the AdamW optimizer with one-cycle learning rate policy, with max learning rate $0.001$, weight decay $0.01$, and momentum $0.85$ to $0.95$. We train the 3D backbone with {the} first decoder layer and FFN for 20 epochs, and the LiDAR-camera fusion components for 6 epochs with batch size {of} 16 using 8 Tesla V100 GPU{s}. We use gradient clipping at an $l_2$ norm of 0.1 to stabilize the training process. The weighting coefficients of heatmap loss, classification loss, and regression loss are $1.0$, $1.0$ and $0.25$, respectively. {The coefficients of matching cost $\lambda_1, \lambda_2, \lambda_3$ are $0.15, 0.25, 0.25$, respectively. The sensitivity analysis of the matching cost coefficients is provided in the next section.}

\noindent\textbf{Waymo.} For Waymo, we only use {a} single sweep as input and set the detection range to $[-75.2m, 75.2m]$ for X and Y {axes}, 
and $[-2m, 4m]$ for Z axis. The maximum number of non-empty voxels is set to 150,000. We adopt the same training strategies as nuScenes except: 1) The first stage training last for 36 epochs with batch size {of} 16 under a max learning rate of $0.001$. 2) The weighting coefficient of regression loss is changed to $2.0$, following CenterPoint. 3) {The matching cost coefficients are set to $0.075, 0.25, 0.25$, respectively.}


\section{Label Assignment Strategy}
\label{supp:label_assign}
Following DETR, we perform label assignment by find{ing} the bipartite matching between predictions and ground\hb{-}truth objects through {the} Hungarian algorithm. In Table~\ref{tab:abl_matching_cost}, we study the effect of the coefficient of each matching cost term on the detection performance of \Name-L. Since the matching results are only decided by the relative value{s of individual matching cost terms}, 
we keep $\lambda_2=0.25$ and try different values for $\lambda_1$ and $\lambda_3$. As shown in Table~\ref{tab:abl_matching_cost}, we find the network suffers from {a} convergence issue when the coefficient of {the} classification cost is too large, and the detection performance is not sensitive to the coefficient within a reasonable range. Since the weighting coefficient{s} of the matching cost {need} 
some tuning, 
we additionally propose a heuristic label assignment strategy~(denoted as \textbf{Heuristic}) to avoid hyper-parameter tuning. {The} \textbf{Heuristic} assignment strategy follows the simple rules: 
each GT box will only be assigned to the predicted box with the same category and {the} smallest center distance. If a conflict appears, the predicted box will be matched to the closer GT box. In this way, we also find the one-to-one matching between prediction and GT but with some GT boxes unused. We find \textbf{Heuristic} works quite well for uncrowded scenes but for objects in a crowded {scene}, such as pedestrian or traffic cone in nuScenes, it is unable to prevent duplicate predictions, which will be further explained in Sec.~\ref{supp:nms}.  

\begin{table}[h]
	\centering
	\resizebox{0.48\textwidth}{!}{
	\begin{tabular}{c|ccc|cc|cc}
		\Xhline{2\arrayrulewidth}
         Matching strategy & $\lambda_1$ & $\lambda_2$ & $\lambda_3$ & {mAP} & {NDS} & {Ped.} & {T.C.} \\
        \hline
         Hungarian & 2.0 & 0.25 & 0.25 & \multicolumn{2}{c|}{Not Converge} & &\\
         Hungarian & 0.5 & 0.25 & 0.25 & 58.5 & 66.0 & 83.8 & 70.5  \\
         Hungarian & 0.25 & 0.25 & 0.25 & 59.2 & 66.1 & 85.2 & 71.6 \\
         Hungarian & 0.15 & 0.25 & 0.25 & 60.0 & 66.8 & 86.1 & 72.1 \\
         Hungarian & 0.1 & 0.25 & 0.25 & 59.3 & 66.3 & 85.3 & 71.6  \\
         \hdashline
         Hungarian & 0.15 & 0.25 & 0.5 & 59.2 & 66.1 & 85.2 & 70.2 \\
         Hungarian & 0.15 & 0.25 & 0.15 & 59.5 & 66.3 & 85.2 & 71.8 \\
         Hungarian & 0.15 & 0.25 & 0.1 &  59.0 & 65.9 & 85.1 & 72.4 \\
         \hdashline
         Heuristic~(w/o~NMS) & & & & 56.7 & 65.3 & 67.3 & 56.6\\
         Heuristic~(w~NMS) & & & & 60.0 & 67.0 & 85.6 & 71.0\\

         \Xhline{2\arrayrulewidth}
	\end{tabular}
	}
	\caption{Ablation study on the matching cost coefficient{s} in Eq.~\ref{eq:matching_cost}. ‘Ped.’, and ‘T.C.’ are short for pedestrian, and traffic cone, respectively. }
	\label{tab:abl_matching_cost}
\end{table}


\section{Effect of NMS}
\label{supp:nms}
Recently, many works~\cite{carion2020endtoend, Sun2020SparseRE, Zhu2021DeformableDD} in 2D detection {have focused} 
on removing the last non-differentiable component, Non-Maximum Suppression~(NMS), in the detection pipeline. OneNet~\cite{Sun2021WhatMF} systematically compares the end-to-end detectors with non-end-to-end detectors, and claims that the \textit{one-to-one positive sample assignment} as well as \textit{classification cost} in the matching cost are the two key factors in producing {a} large score gap between duplicate prediction and building {an} end-to-end detection system without NMS. We refer readers to the original paper~\cite{Sun2021WhatMF} for more details. Following DETR's label assignment strategy, our method naturally satisfies these two requirements and do not need NMS. As show in Table~\ref{tab:nms},  our method still maintains a high mAP without NMS while CenterPoint drops about 12\% mAP. This advantage eliminates the hand-designed NMS post-processing step and makes \Name~more practical and handy for deployment to new scenarios in the real applications. Besides, {since} the \textbf{Heuristic} strategy mentioned in Sec.~\ref{supp:label_assign} does not have classification cost involved in the assignment stage{, this strategy} 
is unable to produce {a} large score gap between duplicate prediction{. This} 
is 
why it does not perform as well as the baseline model on Pedestrian and Traffic cone.


\begin{table}[h]
	\centering
	\resizebox{0.35\textwidth}{!}{
	\begin{tabular}{c|cc}
		\Xhline{2\arrayrulewidth}
         Method & {with~NMS} & {without~NMS} \\
        \hline
         CenterPoint & 57.41 & 45.70 \\
         \Name-L & 59.95 & 59.98 \\
         \Name & 65.58 & 65.60\\

        \Xhline{2\arrayrulewidth}
	\end{tabular}
	}
	\caption{Effect of NMS. We report the mAP of CenterPoint and our \Name~on nuScenes validation set. The results of CenterPoint are reproduced using {MMDetection3D}, 
	which also use{s} a resolution of $(0.075m, 0.075m, 0.2m)$ without any test time augmentation. }
	\label{tab:nms}
\end{table}

\section{Pillar-based 3D Backbone}
\label{supp:pillar}
To demonstrate our framework{'s compatibility} with other 3D backbones, we use PointPillars as our 3D backbone to produce the BEV features{, while keeping} all the other settings the same as the main experiments. The voxel size is set to $(0.2m, 0.2m)$. As shown in Table~\ref{tab:nusc_pillar}, our model outperforms CenterPoint by a remarkable margin under the same pillar-based backbone, showing {its great} generalization {ability}.

\begin{table}[h]
\vspace{-0.2cm}
\setlength{\abovecaptionskip}{0.0cm}
\setlength{\belowcaptionskip}{-0.35cm}
	\centering
	\resizebox{0.35\textwidth}{!}{
	\begin{tabular}{l:cc:cc}
		\Xhline{2\arrayrulewidth}
		& \multicolumn{2}{c:}{\textbf{PointPillars}} & \multicolumn{2}{c}{\textbf{VoxelNet}} \\
          & {mAP} & {NDS} & {mAP} & {NDS} \\
        \hline
        \textbf{CenterPoint} & 50.3 & 60.2 & 59.6 & 66.8\\
        \textbf{\Name-L} & 54.5 & 62.7 & 65.1 & 70.1 \\
        \textbf{\Name} & 58.3 & 64.5 & 67.5 & 71.3\\
        
         \Xhline{2\arrayrulewidth}
	\end{tabular}
	}
	\caption{Results on nuScenes validation set.}
	\label{tab:nusc_pillar}
\end{table}

\section{Discussions of the 2D Network}
\label{supp:2dnet}
Current multi-modality detection models usually employ CNN features from 2D networks pre-trained on different tasks~(i.e., segmentation or detection) and with different resolution~(i.e., different level{s} from ResNet or DLA). There is {no} 
existing work analyzing what kind of image features are most useful for a 3D detection model, 
{and using improper image features}
might prevent the release of the potential for a multi-modality detection system. We believe that the sequential design of our method enables a flexible and off-the-shelf experiment base to explore the effect{s} of different image features. Therefore, we explore this question by fixing the 3D backbone with {the} first decoder layer and performing the second stage training with different image features.

\begin{table}[h]
	\centering
	\resizebox{0.45\textwidth}{!}{
	\begin{tabular}{c|c|cc}
		\Xhline{2\arrayrulewidth}
         Arch. & Task & mAP & NDS \\
        \hline
         DLA34 & Monocular 3D Det. & 65.6 & 69.7 \\
         R50+FPN Level~0 & 2D Det. & 66.4 & 70.1 \\
         R50+FPN Level~0 & 2D Instance Seg. & 66.6 & 70.1\\
         R50+FPN Level~1 & 2D Instance Seg. & 66.5 & 70.1\\
         R50+FPN Level~2 & 2D Instance Seg. & 66.3 & 70.0\\
         R50+FPN Level~3 & 2D Instance Seg. & 65.4 & 69.6\\

         \Xhline{2\arrayrulewidth}
	\end{tabular}
	}
	\caption[STH]{mAP under different 2D backbones. `Det.' and `Seg.' are short for detection and segmentation, respectively. For DLA34 on monocular 3D detection, we acquire the CenterNet pre-trained on nuScenes\footnotemark~following PointAugmenting. For ResNet50 on instance segmentation, we acquire the model pre-trained on nuImages from MMDetection3D. For ResNet50 on 2D detection, we train the model by ourself using MMDetection3D since there is no open-sourced model weights.}
	\label{tab:abl_matching_cost}
\end{table}

{
	From Table~\ref{tab:abl_matching_cost}, we find image features of the 2D instance segmentation model bring the largest performance boost compared with that of detection models. In terms of different levels of the feature pyramid, the feature map of level 0~(stride 4) brings a slightly larger performance gain. We suspect the image features at that level contain more fine-grained information which is important to distinguish small or distant objects. Image features from level 1~(stride 8) and level 2~(stride 16) can bring a similar gain with a smaller resolution of feature maps, while image features from level 3~(stride 32) yields a drop of 1.2\% mAP in comparison with the level-0 counterpart due to the row resolution.
}
\footnotetext{\url{https://github.com/xingyizhou/CenterTrack}}

\section{Adapt Queries at Test Time}
\label{supp:num_query}
Unlike DETR, our object queries are non-parametric and \ourqueryinit. {These two characteristics allow} 
us to use different number{s} of queries during inference. 
{It}
could be useful when we have some prior knowledge about {a} 
scene, such as {its} 
crowdedness. In Table~\ref{tab:num_query}, we provide the performance evaluated under different object queries for the same model trained under $N=200$ queries. Note that we use $N=200$ to get all the number{s} in the main text for its better performance-efficiency trade-off and use $N=300$ for online submission for a slightly better performance.

\begin{table}[h]
	\centering
	\resizebox{0.32\textwidth}{!}{
	\begin{tabular}{c|cccc}
		\hline
         \#queries & 100 & 200 & 300 & 500 \\
        \hline
         mAP & 64.2 & 65.6 & 65.9 & 66.0\\ 
         NDS & 69.2 & 69.7 & 69.8 & 69.8\\
         \hline
	\end{tabular}
	}
	\caption{Results with different number{s} of queries. We keep the model unchanged and only use different number{s} of queries for evaluation.}
	\label{tab:num_query}
\end{table}

\section{Dicussions on Waymo}
\label{supp:waymo_dis}

Our \Name~brings smaller performance gain over \Name-L on Waymo compared with that on nuScenes. {We speculate that this is mainly due to the following two reasons:} 
\begin{enumerate}[itemsep=-1mm]
	\item[(i)] As shown in Table~\ref{tab:nuscene_test}, compared with \Name-L, \Name~brings the largest performance increase for bicycle~(+8.7\%), motorcycle~(+5.4\%), and construction vehicle~(+4.9\%) in terms of mAP on nuScenes. 
	Due to the geometrical ambiguity,
	objects from the above three categories are difficult to distinguish using LiDAR information only, {and} thus the semantic information of images is extraordinarily important for {more} accurate classification. However, the categorization of Waymo is rather coarse-gained~(i.e., vehicle, pedestrian, cyclist), which hides the improvement brought by the image information to some extent.
	\item[(ii)] The LiDAR point clouds in Waymo are much denser than those in nuScenes~(see Sec.~\ref{supp:visualization} for visualizations). Thus the bounding {box} 
	predictions of \Name-L are already with accurate localization, which reduces the room for further improvement {by} 
	image fusion. 
\end{enumerate}

\section{Qualitative Results}
\label{supp:visualization}
We first compare the detection results of \Name~and \Name-L on the nuScenes dataset in Fig.~\ref{fig:vis_L_vs_LC}. The image information improves the performance of the LiDAR-only baseline through reducing the False Positive and False Negative.
More visualization results on Waymo and nuScenes datasets are shown in Fig.~\ref{fig:vis_waymo} and Fig.~\ref{fig:vis_nusc}, respectively.

\begin{figure*}[t]
	\centering
	\includegraphics[width=16cm]{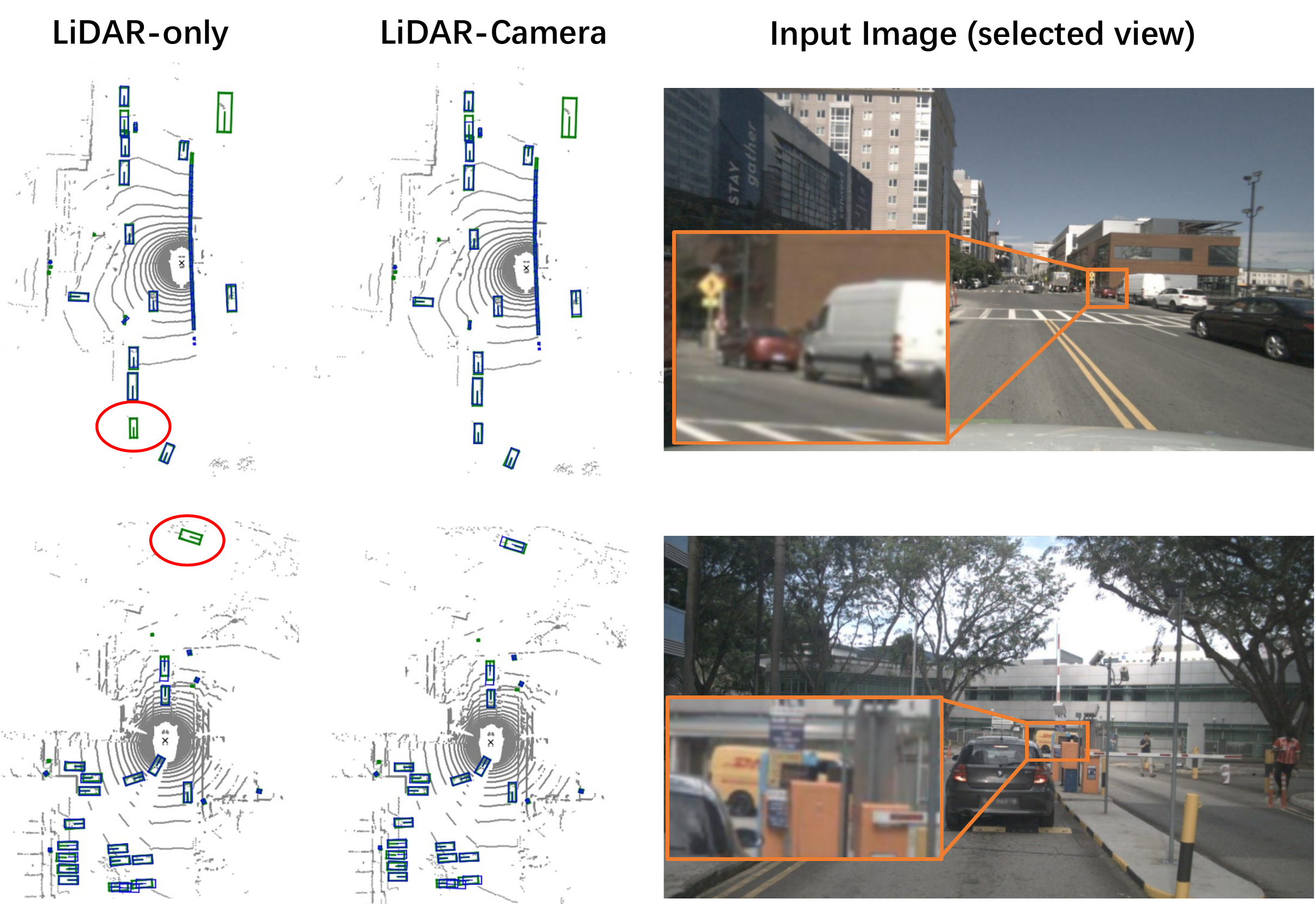}
	\includegraphics[width=16cm]{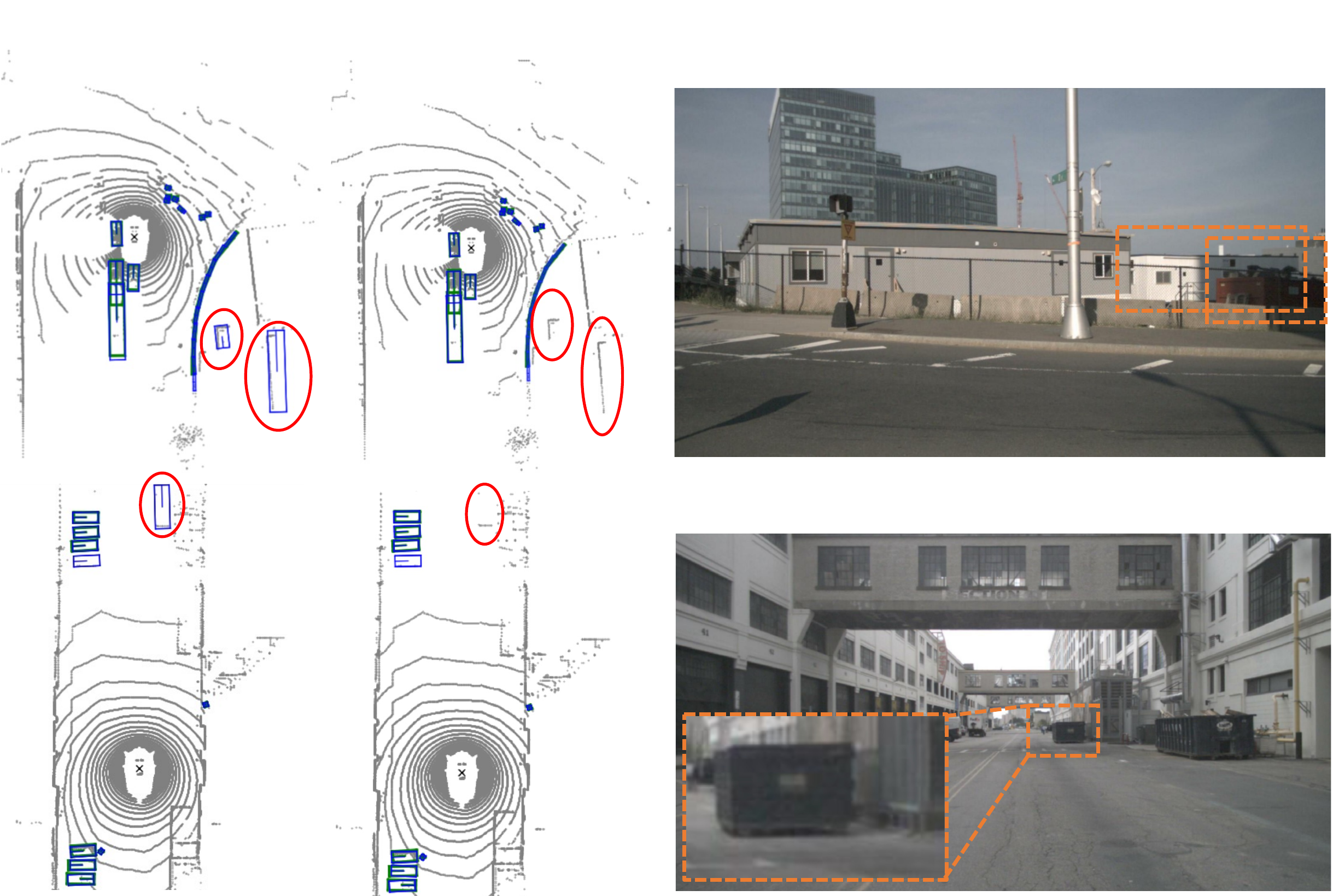}
	\caption{Qualitative comparison between \Name-L and \Name~on the nuScenes dataset. Blue boxes and green boxes are the predictions and ground-truth, respectively. Best viewed with color and zoom-in.}
	\label{fig:vis_L_vs_LC}
\end{figure*}

 \begin{figure*}[ht]
\centering
    \includegraphics[width=18cm]{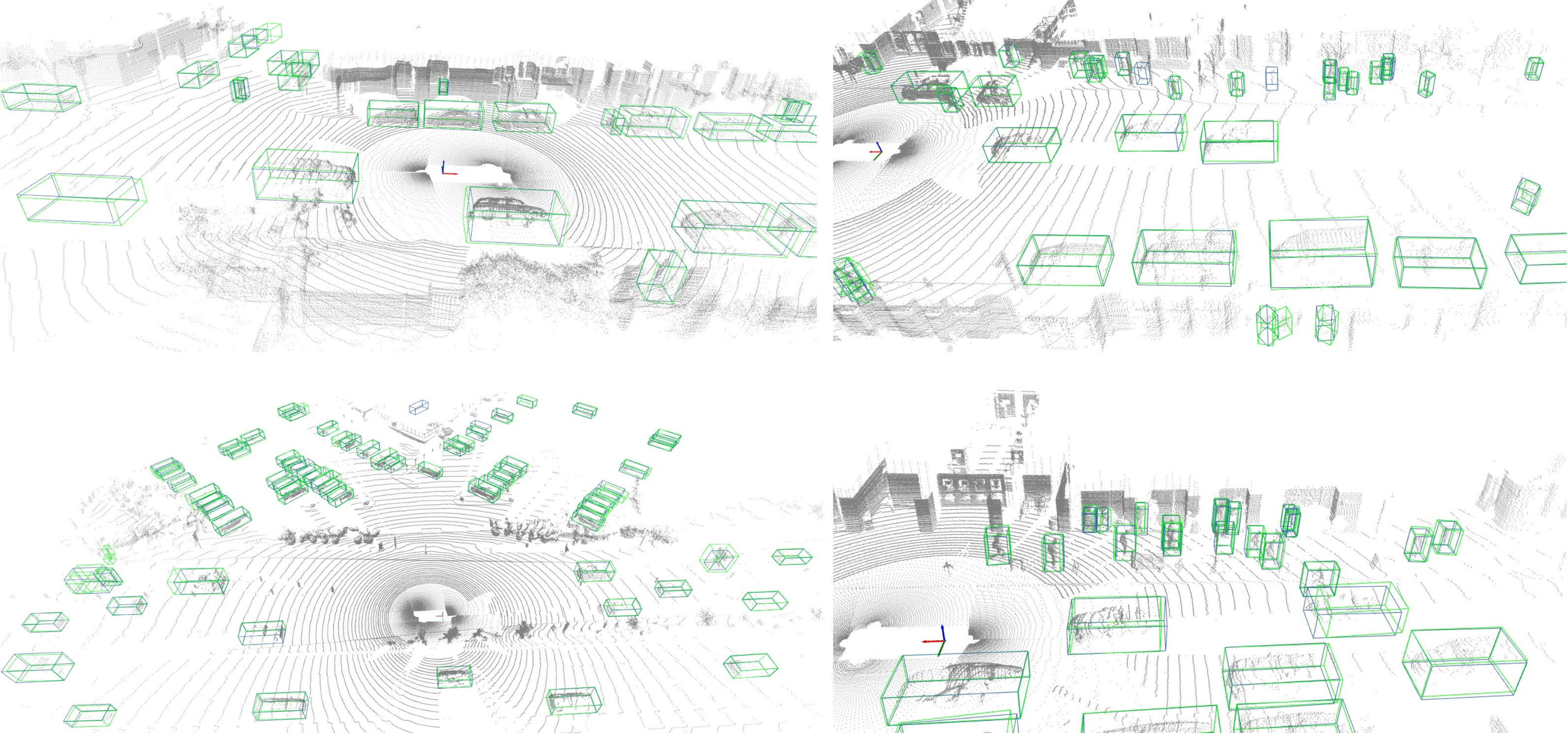}
    \caption{Visualization of detection results on the Waymo dataset. Our model predicts highly accurate bounding boxes for nearby vehicles and pedestrians (note that cyclists are very rare in the dataset) and also handles objects with severe occlusion. Blue boxes and green boxes are the predictions and ground-truth, respectively. Best viewed with color and zoom-in.}
    \label{fig:vis_waymo}
\end{figure*}

 \begin{figure*}[ht]
\centering
    \includegraphics[width=18cm]{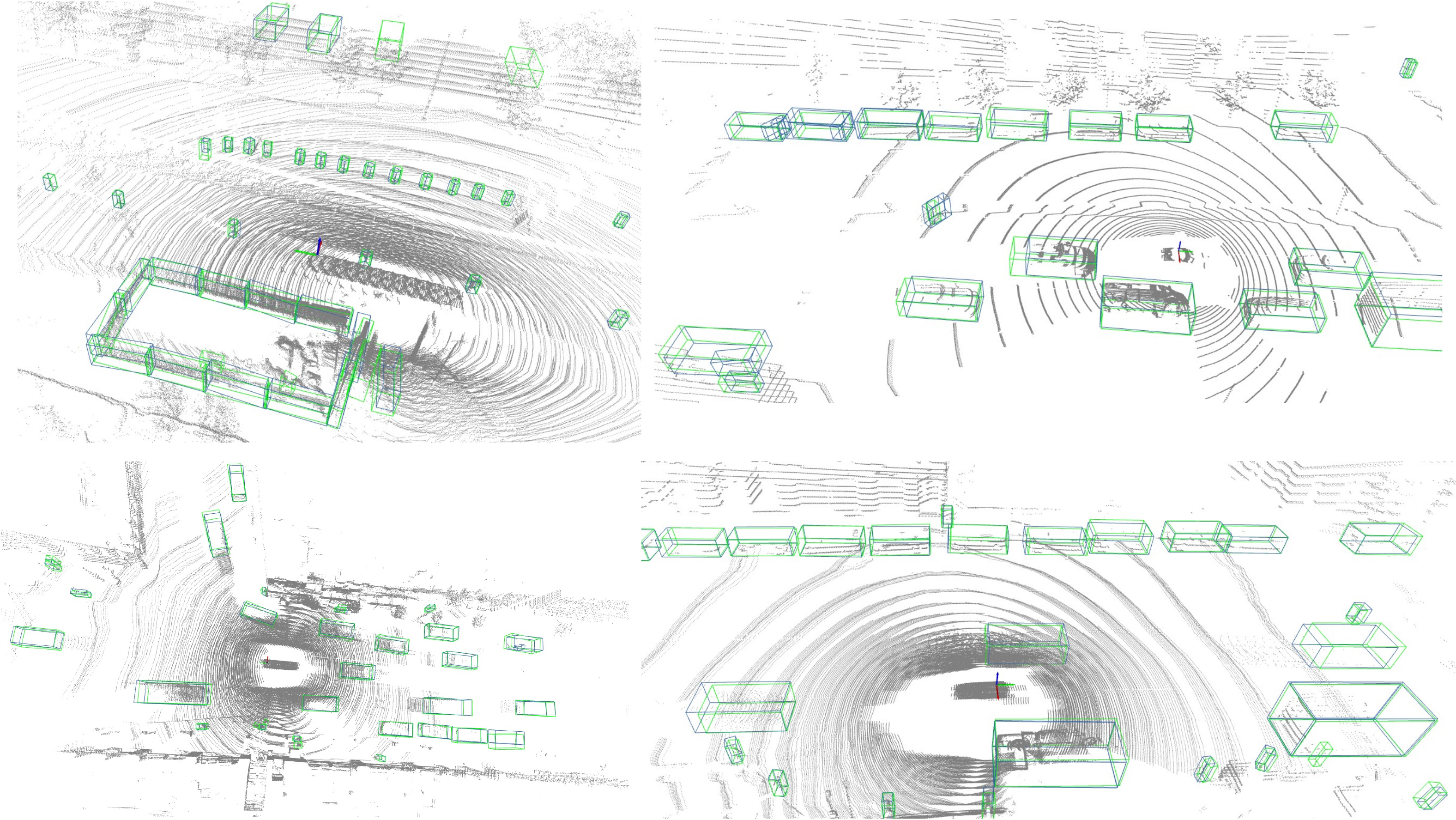}
    \caption{Visualization of detection results on the nuScenes dataset. Compared with Waymo, nuScenes has much {sparser} 
    point clouds and smaller objects such as traffic cones. Nevertheless, our model successfully detects such objects even with only few points observed. Blue boxes and green boxes are the predictions and ground-truth, respectively. Best viewed with color and zoom-in.}
    \label{fig:vis_nusc}
\end{figure*}

}

\end{document}